\documentclass[sigconf,nonacm]{acmart}

\title{Cost-Aware Model Selection for Text Classification: Multi-Objective Trade-offs Between Fine-Tuned Encoders and LLM Prompting in Production}

\author{Alberto Andres Valdes Gonzalez}
\affiliation{
  \institution{Pontifical Catholic University of Chile}
  \city{Santiago}
  \country{Chile}
}
\email{anvaldes@uc.cl}

\usepackage{graphicx}
\usepackage{subcaption}

\newcommand{\subfigwidth}{0.25\textwidth}

\begin{document}

\begin{abstract}
Large language models (LLMs) such as GPT-4o~\cite{openai2024gpt4o} and Claude Sonnet 4.5~\cite{anthropic2025claude45} have demonstrated strong capabilities in open-ended reasoning and generative language tasks, leading to their widespread adoption across a broad range of NLP applications. However, for structured text classification problems with fixed label spaces, model selection is often driven by predictive performance alone, overlooking critical operational constraints encountered in production systems.

In this work, we present a systematic comparison of two contrasting paradigms for text classification: zero- and few-shot, prompt-based large language models, and fully fine-tuned encoder-only architectures. We evaluate these approaches across four canonical benchmarks—IMDB~\cite{maas2011imdb}, SST-2~\cite{socher2013sst}, AG News~\cite{zhang2015agnews}, and DBPedia~\cite{lehmann2015dbpedia}—measuring not only predictive quality (macro F1), but also inference latency and monetary cost.

We frame model evaluation as a multi-objective decision problem and analyze trade-offs using Pareto frontier projections and a parameterized utility function reflecting different deployment regimes. Our results show that fine-tuned encoder-based models from the BERT family achieve competitive, and often superior, classification performance while operating at one to two orders of magnitude lower cost and latency compared to zero- and few-shot LLM prompting. Moreover, their open-source nature enables on-premise deployment, offering additional advantages in privacy, data governance, and reproducibility.

Taken together, our findings suggest that indiscriminate use of large language models for standard text classification workloads can lead to suboptimal system-level outcomes. Instead, fine-tuned encoders emerge as robust and efficient decision engines for structured NLP pipelines, while LLMs are more effectively positioned as complementary components within hybrid architectures. We release all code, datasets, and evaluation protocols to support reproducibility and to establish a reusable benchmark for cost-aware NLP system design.
\end{abstract}


\maketitle

\section{Introduction}

The recent rise of large language models (LLMs), exemplified by systems such as GPT-4o~\cite{openai2024gpt4o} and Claude Sonnet 4.5~\cite{anthropic2025claude45}, has significantly reshaped the landscape of natural language processing (NLP). As general-purpose foundational models~\cite{bommasani2021foundation}, these architectures have demonstrated strong performance across a wide range of open-ended tasks, including reasoning, code generation, and conversational interaction~\cite{bubeck2023sparks,zhao2023llmsurvey}. Their ability to perform tasks via zero-shot and few-shot prompting has further accelerated adoption, fostering the perception that a single, highly expressive model class may suffice for diverse NLP workloads.

Despite this broad success, the applicability of LLMs to structured prediction problems warrants closer examination. In fixed-label text classification—where the objective is to assign each input to one of a predefined set of categories—performance alone is rarely the sole determinant of practical suitability. Such tasks have traditionally been addressed using encoder-only architectures, including BERT~\cite{devlin2019bert}, RoBERTa~\cite{liu2019roberta}, and DistilBERT~\cite{sanh2019distilbert}. When fine-tuned on task-specific data, these models achieve strong and reliable predictive accuracy while maintaining low inference latency and modest computational footprints~\cite{zhao2024advancing,bucher2024smallllms}. Crucially, their open-source nature enables on-premise deployment, offering advantages in privacy preservation, data governance, and auditability that are difficult to guarantee in API-based inference settings.

The growing reliance on LLMs for classification thus raises an important systems-level question: under what conditions does the generality of large, prompt-driven models justify their operational overhead relative to specialized, fine-tuned encoders? Unlike discriminative encoders, LLMs must generate output token by token—even when producing a single class label—introducing inherent latency and usage-based costs. While such overheads may be negligible in low-throughput or exploratory settings, they become consequential at production scale, where classification pipelines may process millions of instances per day under strict latency budgets and cost constraints~\cite{min2024effective,kwon2023vllm}.

In this work, we argue that model selection for text classification should be treated as a multi-objective decision problem rather than a purely accuracy-driven comparison. To this end, we present a systematic empirical study across four canonical English benchmarks—IMDB~\cite{maas2011imdb}, SST-2~\cite{socher2013sst}, AG News~\cite{zhang2015agnews}, and DBPedia~\cite{lehmann2015dbpedia}—comparing zero- and few-shot, prompt-based LLMs against fully fine-tuned encoder-only architectures. Our evaluation jointly considers predictive performance (macro F1), inference latency, and economic cost, capturing the principal dimensions that determine production viability.

Rather than collapsing these metrics into a single score, we analyze their interactions through Pareto frontier projections and a parameterized utility function reflecting different operational priorities. This framework enables us to characterize regimes in which fine-tuned encoders or LLM-based approaches are preferable, depending on accuracy requirements, throughput demands, cost sensitivity, and governance constraints. By grounding model selection in explicit trade-offs, our study reinforces a simple but often overlooked principle: effective NLP systems are built by matching model complexity to problem structure, rather than defaulting to the most expressive available architecture.

\subsection{Model Selection as a Knowledge-Based Process}

In production-grade NLP systems, model selection is rarely a single-objective optimization problem driven solely by predictive accuracy. Instead, it constitutes a knowledge-based decision process in which empirical performance evidence must be evaluated alongside system-level constraints, including latency service-level objectives, throughput requirements, operational budgets, and governance considerations such as reproducibility, auditability, and long-term maintainability. Under these conditions, a model that offers marginal gains in F1 score may be unsuitable if it introduces unstable latency profiles, opaque inference mechanisms, or recurring usage-based costs that scale unfavorably with workload volume.

Viewed through this lens, benchmark results serve not merely as leaderboard rankings, but as reusable decision evidence. The objective of this work is therefore not only to compare encoder-based architectures and prompt-based LLM approaches, but to construct an empirical knowledge base that supports principled model selection under realistic deployment regimes. By jointly quantifying predictive quality, inference latency, and economic cost across representative datasets, we provide the information required to reason about trade-offs in a systematic and transparent manner.

This framing naturally aligns with the design principles of knowledge-based systems, in which decision components are expected to be reliable, interpretable, and controllable over time. In such settings, the choice of a classification model cannot be decoupled from broader architectural concerns, including versioning strategies, monitoring pipelines, and compliance with data governance policies. Moreover, the availability of open-source encoder models enables fully on-premise deployment, offering additional advantages in privacy preservation and operational control that are often critical in regulated environments.

By treating model selection as a knowledge-driven process rather than a one-off performance comparison, this work aims to support the design of scalable, cost-aware, and auditable NLP pipelines. Rather than advocating a single best model, our analysis emphasizes matching model complexity to task structure and operational constraints, thereby reinforcing the principle of using the appropriate model for the problem at hand.

\section{Related Work and Contribution}

Text classification has traditionally been dominated by encoder-only architectures trained under discriminative objectives. Since the introduction of BERT~\cite{devlin2019bert}, Transformer-based encoders such as RoBERTa~\cite{liu2019roberta} and DistilBERT~\cite{sanh2019distilbert} have achieved strong and stable performance across a wide range of sentiment, topic, and intent classification benchmarks. These models are particularly well suited to structured prediction settings, where the output space is finite and well defined, and where task-specific fine-tuning enables efficient learning of decision boundaries with low inference overhead.

In parallel, the emergence of large language models (LLMs) such as GPT-4o~\cite{openai2024gpt4o} and Claude Sonnet 4.5~\cite{anthropic2025claude45} has introduced a fundamentally different paradigm based on instruction following and prompt-based inference~\cite{bommasani2021foundation,bubeck2023sparks}. These models have demonstrated impressive generalization capabilities in open-ended tasks including summarization, question answering, and reasoning. However, their suitability for fixed-label text classification remains an open question. Recent empirical studies~\cite{kostina2025llmsclassification,bucher2024smallllms,chen2025biomedllmbenchmark,zhao2024advancing} indicate that, despite strong average performance, LLM-based classifiers often exhibit higher variance and reduced boundary consistency on standard benchmarks such as IMDB~\cite{maas2011imdb}, SST-2~\cite{socher2013sst}, AG News~\cite{zhang2015agnews}, and DBPedia~\cite{lehmann2015dbpedia}. This behavior is commonly attributed to their generative inference mechanism, which relies on autoregressive token generation rather than direct optimization of discriminative objectives.

A growing body of work has explored the trade-offs between prompt-based learning and task-specific fine-tuning for classification tasks. Prior studies consistently show that specialization through fine-tuning remains critical for achieving robust performance, particularly in domain-specific or efficiency-constrained settings~\cite{luo2023exploring,zhao2024advancing}. These findings suggest that, while instruction-tuned LLMs offer flexibility and ease of use, encoder-based models retain advantages in both predictive stability and computational efficiency for structured classification problems.

Complementary analyses from a systems and deployment perspective further highlight practical limitations of LLM-based classification pipelines. The reliance on autoregressive decoding and remote API-based inference introduces non-trivial latency and recurring operational costs, which can become prohibitive in high-throughput, enterprise-scale environments~\cite{min2024effective,kwon2023vllm}. From this viewpoint, model generality alone is insufficient as a selection criterion; deployment characteristics play a decisive role in determining overall system viability.

Insights from scaling laws and efficiency-oriented analyses~\cite{chung2022scaling,radford2018improving,kaplan2020scaling} further reinforce this perspective, showing that increased model size does not necessarily translate into proportional gains for deterministic classification tasks. Taken together, this line of work converges on a practical principle: the effectiveness of a model is inherently task- and context-dependent.

Despite these advances, existing studies rarely translate empirical findings into actionable guidance for practitioners operating under real-world constraints. In particular, prior work often reports accuracy in isolation, without integrating latency, cost, and governance considerations into a unified decision framework. Our work addresses this gap through three primary contributions:
\begin{itemize}
\item We construct a reproducible empirical knowledge base benchmarking fine-tuned BERT-family encoders and zero- and few-shot LLM prompting across canonical text classification datasets under a unified evaluation protocol, jointly reporting predictive performance (macro F1), inference latency, and monetary cost.
\item We formalize model selection as a knowledge-based, multi-objective decision problem and analyze trade-offs using Pareto frontier projections and a parameterized utility function, enabling explicit reasoning over different deployment regimes and operational priorities.
\item We derive practical implications for production-oriented NLP systems, emphasizing why controllability, stability over time, and governance properties—such as versioning, auditability, and privacy-preserving deployment—should be treated as first-class criteria alongside predictive accuracy.
\end{itemize}

\subsection{Knowledge Artifacts for Operational Decision-Making}

Empirical evaluation in applied machine learning serves a dual role: it advances scientific understanding while simultaneously generating operational knowledge for real-world system design. In production environments, engineering teams repeatedly face similar decision points—whether to rely on hosted LLM APIs, deploy in-house encoder-based models, or adopt hybrid architectures—yet such choices are often made under uncertainty, guided by anecdotal evidence or partial assessments of accuracy, latency, cost, and operational control.

From a knowledge-based systems perspective, a rigorously designed benchmark can be viewed as a persistent \emph{knowledge artifact}. When accompanied by transparent experimental protocols and reproducible implementations, such artifacts transcend one-off empirical comparisons and instead encode reusable decision knowledge. They enable systematic reasoning about architectural trade-offs across organizations, domains, and evolving infrastructure contexts, remaining valuable as models, providers, and deployment constraints change over time.

In line with this perspective, we structure our study around deployment-relevant decision variables rather than accuracy in isolation. Performance metrics are analyzed jointly with inference latency and monetary cost through Pareto frontier projections and a parameterized utility function, allowing practitioners to reason explicitly about trade-offs under different operational priorities. The resulting evaluation tables, scripts, and procedures are intentionally designed for reuse: pricing assumptions can be updated, hardware profiles substituted, and experiments re-executed as systems and constraints evolve.

By framing benchmarking as a knowledge-driven process rather than a static academic exercise, this work supports responsible and auditable decision-making in production NLP systems. The released artifacts are intended to function as living reference points that facilitate sustainable system evolution, enabling teams to continuously reassess model choices as performance requirements, cost structures, and governance constraints change.

\section{Method: Fine-Tuned Encoders vs. Zero-/Few-Shot LLM Prompting}
\label{sec:method}

\subsection{Operational Assumptions and Decision Variables}

To ensure that model evaluation reflects real-world deployment considerations, we explicitly ground our methodology in a set of operational assumptions that commonly constrain production NLP systems. Rather than treating model selection as an abstract optimization problem, we frame it as a decision process shaped by recurring system-level requirements encountered in practice.

We consider three primary operational constraints. First, \emph{latency budgets}, defined as end-to-end per-request service-level objectives that bound acceptable response times. Second, \emph{throughput requirements}, expressed in requests per second or samples processed per day, which determine the scalability demands placed on the inference stack. Third, \emph{budget constraints}, capturing the recurring monetary cost of inference at production scale rather than one-time training expenses. Together, these constraints define the feasible operating regimes for text classification systems deployed in production environments.

In addition to these quantitative constraints, we incorporate governance-related decision variables that frequently play a decisive role in enterprise and regulated settings. These include \emph{reproducibility}, \emph{auditability}, and \emph{operational controllability}. Such properties encompass the ability to fix and version model behavior, reproduce historical predictions, monitor drift, and perform safe rollbacks over time. These considerations are particularly salient when model decisions have downstream business, legal, or compliance implications.

Our evaluation metrics are selected to map directly onto these decision variables. Predictive reliability is measured using macro F1, reflecting balanced performance across classes. Operational responsiveness is quantified through end-to-end inference latency and time-to-first-token (TTFT), reported using percentile statistics (p50, p95, and p99) to capture both typical and tail behavior under realistic load conditions. Economic feasibility is assessed via snapshot-based inference cost estimates, corresponding to contemporaneous pricing models and avoiding amortized or hypothetical cost assumptions.

By elevating latency, TTFT, and monetary cost to first-class evaluation signals—rather than treating them as secondary engineering concerns—our methodology is designed to function as actionable decision evidence. This framing enables systematic reasoning about trade-offs between fine-tuned encoders and prompt-based LLM approaches, directly supporting the design, deployment, and governance of cost-aware, knowledge-based NLP systems.

\subsection{Reproducibility as a Knowledge Constraint}

In knowledge-based systems, a model is not merely a predictive component, but a foundational element of an auditable and evolving decision pipeline. Fine-tuned encoder-based models naturally yield stable and inspectable artifacts, including trained weights, tokenizers, configuration files, and training logs. These artifacts can be versioned, reproduced, calibrated, and rolled back, enabling traceable system behavior and controlled updates in response to data drift or changing requirements. Such an artifact-centric workflow supports systematic validation of decision logic and long-term stability in production environments.

In contrast, API-based deployments of large language models introduce external dependencies that may weaken knowledge reliability. Inference behavior can change as a result of provider-side model updates, pricing structures may evolve independently of system design, and latency characteristics depend on cloud infrastructure, network conditions, and service availability that remain outside organizational control~\cite{weidinger2022taxonomy,dodge2022measuring}. While API-based LLMs are highly effective for rapid prototyping and reasoning-intensive tasks~\cite{brown2020gpt3,touvron2023llama2}, these sources of variability become critical when deterministic labeling, stable decision boundaries, and reproducible behavior at scale are required.

An additional limitation arises from the restricted access to calibrated predictive signals in many API-based inference settings. Fine-tuned encoder models expose logits and class-level probability estimates, enabling post-hoc calibration, uncertainty-aware thresholds, and consistent confidence scoring. By contrast, LLM APIs typically return discrete outputs or provider-specific likelihood proxies that are not directly comparable across model versions or providers, constraining their applicability in systems that require probabilistic interpretability and stable confidence estimation.

For these reasons, we treat reproducibility, controllability, and probabilistic transparency as first-class constraints in model selection, alongside accuracy, latency, and monetary cost. This perspective reflects the requirements of production-oriented knowledge systems, in which decision consistency, auditability, and uncertainty management are essential to maintaining trust, compliance, and long-term operational viability.

\subsection{Tasks and Datasets}

We frame text classification as a fixed-label supervised learning problem and evaluate across four canonical English benchmarks: \textbf{IMDB} (binary sentiment) \cite{maas2011imdb}, \textbf{SST-2} (binary phrase-level sentiment) \cite{socher2013sst}, \textbf{AG News} (4-class topic classification) \cite{zhang2015agnews}, and \textbf{DBPedia} (14-class ontology classification) \cite{lehmann2015dbpedia}. These datasets span varying label cardinalities, dataset sizes, and domain characteristics, providing a representative testbed for structured text classification.

For each dataset, we adopt the official splits when available and ensure that all train/validation/test partitions are constructed using stratified sampling, preserving the original class distributions across splits. This design choice mitigates class imbalance effects and ensures comparability across model families and experimental runs.

We report the effective train/validation/test sizes used in our experiments for each dataset in Table~\ref{tab:datasets}, consistent with the evaluation protocol adopted throughout this work.

\begin{table}[h]
\centering
\caption{Datasets used in the evaluation and effective split sizes.}
\label{tab:datasets}
\begin{tabular}{lcccc}
\toprule
\textbf{Dataset} & \textbf{Task} & \textbf{\#Classes} & \textbf{Train} & \textbf{Val / Test} \\
\midrule
IMDB     & Sentiment        & 2  & 25{,}000  & 12{,}500 / 12{,}500 \\
SST-2    & Sentiment        & 2  & 47{,}144  & 10{,}102 / 10{,}103 \\
AG News  & Topic            & 4  & 120{,}000 & 3{,}800 / 3{,}800   \\
DBPedia  & Ontology         & 14 & 560{,}000 & 63{,}000 / 7{,}000  \\
\bottomrule
\end{tabular}
\end{table}

\subsection{Model Families}

\paragraph{Encoders (fine-tuning).}
We consider encoder-only Transformer architectures representative of the BERT family: \textbf{BERT} \cite{devlin2019bert}, \textbf{RoBERTa} \cite{liu2019roberta}, and \textbf{DistilBERT} \cite{sanh2019distilbert}. For each dataset, we attach a standard linear classification head on top of the pooled \texttt{[CLS]} representation and fine-tune the model end-to-end using supervised training.

\paragraph{LLMs (prompting).}
We evaluate instruction-following LLMs in \textit{zero-shot} and \textit{few-shot} prompting settings, without gradient updates: GPT-4o \cite{openai2024gpt4o}, and Claude Sonnet 4.5~\cite{anthropic2025claude45}, alongside related large-scale models discussed in the literature \cite{brown2020gpt3,touvron2023llama2}. Prompts explicitly specify the label set and enforce a constrained output format. Few-shot prompts include \(k\) in-context exemplars per class, sampled from the training set and kept fixed across runs to ensure comparability.

\subsection{Experimental Environment and Deployment Setup}

All experiments are conducted in a controlled cloud-based environment designed to reflect realistic production deployment conditions. Model training and evaluation workflows are orchestrated via Jupyter notebooks executed on Google Colab. Fine-tuning of encoder-based models is performed using NVIDIA A100 GPUs, while API-based LLM evaluations are executed on CPU runtimes, reflecting typical inference usage patterns.

Encoder models are containerized and deployed as stateless inference services on Google Cloud Run, with container images stored in Google Artifact Registry. This setup enables consistent measurement of end-to-end inference latency and cost, capturing the full request--response lifecycle under sustained utilization. LLM-based models are accessed exclusively via provider-hosted APIs, without local deployment or gradient updates.

This deployment configuration reflects a common industry-grade architecture and ensures that latency and cost measurements correspond to realistic operational settings rather than isolated hardware benchmarks.

Deployment configurations, container specifications, and Cloud Run service definitions for encoder-based models are released at \url{https://github.com/anvaldes/bert-family-deploy}.

\subsection{Training and Inference Protocol}
\label{sec:training_inference}

\paragraph{Fine-tuning schedule.}
Encoder models are fine-tuned for up to four epochs, with optimal validation performance typically reached within the first one epoch. To enable epoch-wise evaluation and checkpointing, training is implemented as a sequence of single-epoch runs (\texttt{num\_train\_epochs=1}), iteratively executing the training loop and evaluating performance after each epoch. Optimization is performed using AdamW \cite{loshchilov2019adamw} with weight decay.

After each epoch, we compute validation macro-F1 and store a full checkpoint, including model weights and tokenizer state. The checkpoint used for final test evaluation is selected offline as the epoch achieving the highest validation macro-F1, strictly avoiding any test-set peeking. Batch size, learning rate, and maximum sequence length are held constant within each dataset–model family to ensure comparability. All hyperparameters and configurations are released to support full reproducibility.

\paragraph{Generalization-aware selection metric.}
To balance predictive performance and generalization stability, checkpoint selection is guided by a generalization-aware score:
\[
\textsf{Score} \;=\; F1_{\text{val}} \;-\; \bigl| F1_{\text{val}} - F1_{\text{train}} \bigr|.
\]
This criterion reduces to \(F1_{\text{val}}\) when the train–validation gap is zero, while penalizing both overfitting (\(F1_{\text{train}} \gg F1_{\text{val}}\)) and anomalous underfitting. The formulation follows classical early-stopping principles and generalization-gap analysis \cite{prechelt1998early,ying2019overfit,powers2011evaluation}.

\paragraph{LLM prompting and decoding.}
LLMs are evaluated using fully deterministic decoding, with temperature set to \(T = 0.0\) and nucleus sampling disabled (\(\textit{top\_p} = 1.0\)). This configuration eliminates stochastic variation in token generation, ensuring reproducible and stable outputs across runs for identical inputs. Output formats are explicitly constrained to map directly to a single class label, minimizing post-processing ambiguity and reducing variance induced by generative decoding.

In few-shot settings, in-context exemplars are drawn from the training set and balanced across classes whenever feasible. Exemplars are fixed across runs to ensure strict comparability. For each request, we report both time-to-first-token (TTFT) and end-to-end inference latency as returned by the API, enabling fine-grained assessment of responsiveness under realistic deployment conditions.

\paragraph{Latency and cost.}
For encoder-based models, we measure end-to-end inference latency after deployment behind a lightweight inference API, capturing the full request–response cycle from client invocation to label output. This setup reflects realistic production usage, including request serialization, model execution, and response handling, rather than isolated hardware-level timing. For LLMs, we report both time-to-first-token (TTFT) and total response latency as provided by the corresponding APIs.

Monetary cost is computed using provider-published pricing snapshots at the time of experimentation. For LLMs, costs are calculated based on aggregated input and output token usage over the evaluation set. For encoder models, inference cost is estimated from the underlying serving infrastructure under sustained utilization consistent with the measured throughput. This unified methodology enables a fair, deployment-oriented comparison of latency and cost across model families.

Explicit cost formulas and token accounting procedures are provided in Appendix.

\paragraph{Repeated runs and random seeds.}
To account for stochastic variability, encoder-based models are fine-tuned using three distinct random seeds for each dataset--model configuration. For LLM-based evaluations, each zero-shot and few-shot configuration is executed three times per dataset, even under deterministic decoding settings, to capture residual variability arising from infrastructure and API-level factors.

All reported results correspond to the mean and standard deviation across runs. This reporting choice reflects deployment-level variability rather than population-level statistical inference.

\subsection{Evaluation Metrics and Significance}

\paragraph{Primary metrics.}
We adopt macro-averaged F1 as the primary evaluation metric, as it provides a class-balanced assessment in multi-class and potentially imbalanced settings \cite{powers2011evaluation}. For completeness, we additionally report precision, recall, and accuracy on both train and validation splits. Model selection is performed exclusively on the validation set using the generalization-aware score defined in Section~\ref{sec:training_inference}, while final test results correspond to the checkpoint achieving the highest validation score, with no test-set exposure during training or selection.

Beyond predictive quality, we report deployment-relevant operational metrics, including end-to-end inference latency, time-to-first-token (TTFT) for LLMs, input and output token counts, and total inference cost. These metrics are treated as first-class signals to support cost- and latency-aware model selection in production-oriented NLP systems.

\paragraph{Robustness checks.}
To reduce variance induced by prompt sensitivity in LLM evaluations, we employ a fixed prompt template and a stable set of in-context exemplars for all zero-shot and few-shot experiments. 

For encoder-based models, we report training time per epoch and explicitly monitor the train--validation performance gap. Checkpoint selection penalizes excessive divergence between training and validation performance via the proposed generalization-aware metric, thereby discouraging overfitting and promoting stable generalization behavior.

\subsection{Metric Computation Details}

Latency metrics are reported as the median of per-seed percentile measurements (p50, p95, and p99) after an initial warm-up phase, ensuring that cold-start effects do not bias results. For encoder-based models deployed on Cloud Run, latency is measured end-to-end, from client request initiation to label response.

Inference cost for encoder models is estimated using p50 latency under sustained utilization and snapshot cloud pricing for the corresponding Cloud Run configuration. For LLM-based models, cost is computed from aggregated input and output token counts over the evaluation set, using provider-published per-token pricing.

Pricing snapshots correspond to publicly available rates as of January 22, 2026. Token statistics, latency measurements, and cost computations are released alongside the evaluation scripts to support full reproducibility.

All reported metrics are aggregated across repeated runs and random seeds, and are presented as mean $\pm$ standard deviation. This reporting choice is intended to characterize deployment-level variability across executions rather than to estimate population-level confidence intervals.

\subsection{Why a Gap-penalized Selection Metric?}

Selecting checkpoints solely by validation F1 can favor models that benefit from stochastic variance or transient optimization effects, rather than robust generalization. To mitigate this risk, we explicitly penalize the absolute train--validation performance gap, operationalizing classical bias--variance trade-offs and early-stopping principles~\cite{prechelt1998early,ying2019overfit}. Models that overfit the training data typically exhibit inflated training scores without commensurate validation improvements, signaling brittle decision behavior.

By incorporating the train--validation gap into the selection criterion, our metric prioritizes checkpoints that combine strong validation performance with stable generalization, rather than maximizing validation F1 in isolation. This selection strategy is applied uniformly across all model families and is used exclusively for checkpoint selection within a given model and dataset, not for performance comparison across paradigms.

From a knowledge-based systems perspective, this emphasis on generalization stability is essential. In production settings, a classifier must function not only as an accurate predictor, but as a reliable and auditable decision component whose behavior remains consistent under deployment noise and moderate distributional shifts. Our empirical results suggest that this selection strategy favors models exhibiting stable decision boundaries under operational constraints, aligning with prior findings in structured text classification~\cite{bucher2024smallllms,zhao2024advancing,chen2025biomedllmbenchmark}.

\subsection{Utility Function and Pareto-based Model Selection}

To formalize model selection under competing operational constraints, we define a utility function that jointly captures predictive quality, inference cost, and latency:
\[
U(F1, Cost, Latency) = \frac{F1}{Cost} \cdot \exp\left(-\frac{Latency_{50}}{\tau}\right),
\]
where $\tau$ represents a latency tolerance parameter controlling the relative importance of responsiveness.

We evaluate this utility function under multiple values of $\tau$, with $\tau = 500$ ms serving as a representative production-scale default. Sensitivity analyses for alternative $\tau$ values are reported to assess robustness under different latency regimes.

In parallel, we apply Pareto dominance criteria to identify non-dominated models across F1, latency, and cost dimensions. A model is considered Pareto-optimal if no other model simultaneously achieves higher F1 while incurring lower latency and cost. Two-dimensional Pareto projections are used to visualize trade-offs under different operational priorities.

All experimental notebooks, evaluation scripts, prompt templates, and metric computation code are publicly available at \url{https://github.com/anvaldes/Finetuned-encoders-LLM-Prompting}.

\section{Results}
\label{sec:results}

\subsection{Overview of Experimental Findings}
\label{sec:results_overview}

We compare fine-tuned encoder-only architectures like BERT~\cite{devlin2019bert},
RoBERTa~\cite{liu2019roberta}, and DistilBERT~\cite{sanh2019distilbert} against instruction-tuned LLM prompting like GPT-4o~\cite{openai2024gpt4o} and Claude Sonnet 4.5~\cite{anthropic2025claude45} on four fixed-label text classification benchmarks: IMDB~\cite{maas2011imdb}, SST-2~\cite{socher2013sst}, AG News~\cite{zhang2015agnews}, and DBPedia~\cite{lehmann2015dbpedia}. Encoder models are fine-tuned and deployed as stateless inference services on Google Cloud Run (container images stored in Artifact Registry), enabling end-to-end latency measurement under a production-like request--response setup. LLM evaluations are conducted via official provider APIs under deterministic decoding ($T=0.0$) and constrained-output prompts, reporting both time-to-first-token (TTFT) and total latency.

All experiments follow the repeated-run protocol defined in Section~\ref{sec:training_inference}: encoder models are fine-tuned using three random seeds per dataset--model pair, while each LLM configuration (zero-shot and few-shot) is executed three times per dataset to capture residual API-level variability. Predictive metrics are reported as mean $\pm$ standard deviation across runs. Latency metrics are reported via percentile statistics (p50/p95/p99) following the median-of-per-seed-percentiles aggregation described in Section~\ref{sec:method}.

Across datasets, we observe a consistent pattern: fine-tuned encoders match or exceed LLM performance on structured classification while delivering substantially lower inference latency and cost. In particular, encoder deployments preserve controllability (versioned artifacts, reproducibility, and on-prem feasibility), whereas API-based LLM inference introduces higher tail latency and markedly higher usage-based costs. To make these trade-offs operational, we analyze model choice as a multi-objective decision problem using (i) a utility-based ranking under multiple latency tolerance regimes (Section~\ref{sec:utility_selection}) and (ii) Pareto frontier projections.

\subsection{Per-dataset Results}
\label{sec:results_tables}

Tables~\ref{tab:imdb}--\ref{tab:dbpedia} report the full set of predictive, latency, token, and cost metrics for each dataset. For brevity, our interpretation focuses on the decision-relevant dimensions: (a) macro-F1, (b) p50/p95 latency and TTFT, and (c) estimated cost at scale under pricing snapshots as of January~22,~2026.

\subsubsection{IMDB Sentiment Analysis}
\label{sec:results_imdb}
Table~\ref{tab:imdb} confirms a familiar but decision-critical regime: on a high-signal binary task like IMDB, LLM prompting achieves only \emph{marginal} predictive gains over specialized encoders, while incurring substantially worse tail latency and usage-based cost. The best encoder (RoBERTa, $94.84\pm0.12$ macro-F1) trails the best LLM configuration by less than two F1 points (Claude 4.5 FS, $96.48\pm0.01$), yet remains markedly more deployable in a production setting: encoder p50 latency remains below one second across all models (234--622\,ms) with comparatively tighter tails. By contrast, both LLM families exhibit higher TTFT and significantly heavier tail latency, with p95 TTFT in the 1.7--1.9\,s range for Claude and p95 end-to-end latency approaching 2\,s, increasing the risk of SLA violations under bursty or high-throughput workloads.

From a cost perspective, the trade-off is decisive: the estimated cost at 1M requests is two orders of magnitude lower for encoders (e.g., DistilBERT: \$12.44; RoBERTa: \$32.98) than for LLM prompting (GPT-4o ZS: \$842.78; Claude 4.5 ZS: \$1174.95). Importantly, few-shot prompting does \emph{not} materially improve predictive performance on IMDB (e.g., GPT-4o: $96.11\rightarrow96.16$; Claude: $96.45\rightarrow96.48$), but roughly doubles average input tokens (GPT-4o: 333$\rightarrow$611; Claude: 367$\rightarrow$682), translating into large cost increases and worse latency tails. Overall, IMDB illustrates that for fixed-label sentiment at scale, encoder fine-tuning remains the dominant choice once cost and tail latency are treated as first-class constraints.

\subsubsection{Stanford Sentiment Treebank v2 (SST-2)}
\label{sec:results_sst2}
Table~\ref{tab:sst2} highlights a different but equally important regime: fine-tuned encoders match or exceed LLM prompting while offering substantially lower latency (roughly 3--10$\times$ at p50, depending on the provider) and about two orders of magnitude lower cost at scale. BERT attains $94.42\pm0.10$ macro-F1, essentially matching Claude 4.5 few-shot ($94.41\pm0.06$), and surpassing GPT-4o even under few-shot ($90.45\pm0.03$). In this dataset, specialization is particularly effective: encoder p50 latency is $\approx$98--147\,ms (p95 $\approx$124--207\,ms), versus LLM p50 latency of 326--1394\,ms and p95 of 519--2048\,ms. TTFT further separates the regimes: LLM p50 TTFT is 323--1328\,ms and p95 TTFT is 516--1944\,ms, making the end-to-end response time largely dominated by network/API overhead rather than computation.

Few-shot prompting is again expensive relative to its benefit profile. For GPT-4o, few-shot increases average input tokens from 72.99 to 150.99 while improving macro-F1 by $\approx$3.5 points; for Claude, few-shot improves macro-F1 by $\approx$2.6 points but increases tokens from 83.89 to 174.89 and raises estimated cost from \$326.67 to \$599.67 per 1M requests. The central lesson is that even when few-shot gains are real, they remain a \emph{budgeted} choice: they must be justified under explicit cost--latency constraints rather than adopted as a default. In contrast, encoders deliver near-ceiling accuracy with stable low-latency behavior, making them the safer deployment baseline for binary sentiment with strict SLAs.

\subsubsection{AG News Topic Classification}
\label{sec:results_agnews}
Table~\ref{tab:agnews} exhibits the clearest specialization advantage for encoder fine-tuning. All three encoders cluster around $\approx94$--$95$ macro-F1 (best: RoBERTa $94.63\pm0.14$), whereas LLM prompting lags substantially even with few-shot (best LLM: Claude 4.5 ZS $91.35\pm0.11$; GPT-4o FS $89.65\pm0.13$). This gap is operationally meaningful: it reflects systematic misalignment between generative prompting and fixed-label multi-class topical boundaries, rather than noise-level variability. Latency further reinforces the production preference for encoders: encoder p50 latency is 108--197\,ms (p95 161--286\,ms), while LLM p50 latency is 332--1435\,ms with heavy p95 tails (582--2544\,ms) and multi-second p99 tails (up to 5.76\,s). TTFT trends similarly, indicating that end-to-end responsiveness in API inference is largely constrained by service overhead.

Cost at scale strongly favors encoders: \$5.73--\$10.44 per 1M requests versus \$276.00--\$1271.58 for LLM prompting. Moreover, few-shot prompting increases average input tokens sharply (e.g., GPT-4o: 106$\rightarrow$357; Claude: 122$\rightarrow$399) without closing the predictive gap, implying an unfavorable cost--accuracy exchange. Overall, AG News strengthens the central production claim of this work: for structured multi-class classification, encoder specialization yields superior accuracy while simultaneously improving latency and cost, making prompting a dominated strategy under realistic throughput requirements.

\subsubsection{DBPedia Ontology Classification}
\label{sec:results_dbpedia}
Table~\ref{tab:dbpedia} is near-saturated by fine-tuned encoders, with BERT and DistilBERT achieving $99.40$ macro-F1 (std $\leq 0.04$), leaving little headroom for alternative approaches. In contrast, LLM prompting remains competitive but consistently below the encoder ceiling (best LLM: Claude 4.5 ZS $98.83\pm0.04$), and the gap is coupled with significantly worse latency and cost. Encoder p50 latency remains 133--206\,ms (p95 217--381\,ms), whereas LLM p50 latency is 406--1128\,ms (p95 610--1863\,ms) with TTFT p50 of 404--1058\,ms and p95 of 609--1775\,ms. Thus, even when LLM accuracy is high, API overhead makes its tail latency materially less predictable.

Cost amplifies the dominance of encoders: \$7.03--\$10.90 per 1M requests versus \$463.20--\$2701.89 for LLM prompting. Few-shot prompting increases average input tokens dramatically (GPT-4o: 181$\rightarrow$757; Claude: 226$\rightarrow$876) yet does not surpass the best encoder, producing a strictly worse cost--utility profile under fixed-label constraints. Taken together, DBPedia provides strong empirical support for a diminishing-returns interpretation: for deterministic ontology-style classification where the label space is fixed and learnable, scaling to large generative models does not translate into better outcomes, while substantially increasing latency uncertainty and operating cost.

For completeness, we collect the full metric tables at the end of this subsection.
All metrics are reported on the held-out test split; validation data is used only for checkpoint selection and is never used for model comparison.

\clearpage

\begin{table*}[t]
\caption{IMDB — Binary sentiment classification results (mean $\pm$ std over three independent repetitions; three seeds for encoders, three API runs for LLM prompting).}
\label{tab:imdb}
\centering
\resizebox{\textwidth}{!}{
\begin{tabular}{lccccccc}
\toprule
\textbf{Metric} &
\textbf{BERT} &
\textbf{DistilBERT} &
\textbf{RoBERTa} &
\textbf{GPT-4o ZS} &
\textbf{GPT-4o FS} &
\textbf{Claude 4.5 ZS} &
\textbf{Claude 4.5 FS} \\
\midrule
F1 Score (\%) & $93.43\pm0.21$ & $92.73\pm0.08$ & $94.84\pm0.12$ & $96.11\pm0.06$ & $96.16\pm0.02$ & $96.45\pm0.01$ & $96.48\pm0.01$ \\
Precision (\%) & $93.49\pm0.18$ & $92.77\pm0.08$ & $94.86\pm0.11$ & $96.13\pm0.06$ & $96.18\pm0.02$ & $96.49\pm0.01$ & $96.53\pm0.01$ \\
Recall (\%) & $93.43\pm0.21$ & $92.74\pm0.08$ & $94.84\pm0.12$ & $96.11\pm0.06$ & $96.16\pm0.02$ & $96.45\pm0.01$ & $96.48\pm0.01$ \\
Accuracy (\%) & $93.43\pm0.21$ & $92.74\pm0.08$ & $94.84\pm0.12$ & $96.11\pm0.06$ & $96.16\pm0.02$ & $96.45\pm0.01$ & $96.48\pm0.01$ \\
\midrule
Inference Latency p50 (ms) & 480.06 & 234.82 & 622.21 & 344.89 & 394.84 & 1312.94 & 1314.52 \\
Inference Latency p95 (ms) & 1121.54 & 519.28 & 1467.37 & 546.43 & 634.90 & 1868.71 & 1952.43 \\
Inference Latency p99 (ms) & 1206.96 & 655.73 & 1699.31 & 900.60 & 1239.38 & 2568.49 & 2509.75 \\
\midrule
TTFT p50 (ms) & N/A & N/A & N/A & 342.33 & 392.35 & 1261.18 & 1261.85 \\
TTFT p95 (ms) & N/A & N/A & N/A & 544.15 & 632.82 & 1723.20 & 1859.10 \\
TTFT p99 (ms) & N/A & N/A & N/A & 898.01 & 1238.56 & 2383.10 & 2380.11 \\
\midrule
Avg. Input Tokens / Req & N/A & N/A & N/A & 333.11 & 611.11 & 366.65 & 681.67 \\
Avg. Output Tokens / Req & N/A & N/A & N/A & 1.00 & 1.00 & 5.00 & 5.00 \\
Warm-up Iterations & 10 & 10 & 10 & 10 & 10 & 10 & 10 \\
Sampling Temperature & N/A & N/A & N/A & 0.00 & 0.00 & 0.00 & 0.00 \\
Top-p Sampling & N/A & N/A & N/A & 1.00 & 1.00 & N/A & N/A \\
\midrule
Training / Inference Setup & FT & FT & FT & ZS & FS & ZS & FS \\
\midrule
Input Token Price (USD / 1M) & N/A & N/A & N/A & 2.50 & 2.50 & 3.00 & 3.00 \\
Output Token Price (USD / 1M) & N/A & N/A & N/A & 10.00 & 10.00 & 15.00 & 15.00 \\
Allocated CPU (vCPU) & 2 & 2 & 2 & N/A & N/A & N/A & N/A \\
Allocated Memory (GiB) & 2 & 2 & 2 & N/A & N/A & N/A & N/A \\
CPU Price (USD / 1M vCPU-s) & 24.00 & 24.00 & 24.00 & N/A & N/A & N/A & N/A \\
Memory Price (USD / 1M GiB-s) & 2.50 & 2.50 & 2.50 & N/A & N/A & N/A & N/A \\
\midrule
Estimated Cost (USD / 1M req) & 25.44 & 12.44 & 32.98 & 842.78 & 1537.78 & 1174.95 & 2120.01 \\
\bottomrule
\end{tabular}}
\end{table*}

\begin{table*}[t]
\caption{SST-2 — Binary sentiment classification results (mean $\pm$ std over three independent repetitions; three seeds for encoders, three API runs for LLM prompting).}
\label{tab:sst2}
\centering
\resizebox{\textwidth}{!}{
\begin{tabular}{lccccccc}
\toprule
\textbf{Metric} &
\textbf{BERT} &
\textbf{DistilBERT} &
\textbf{RoBERTa} &
\textbf{GPT-4o ZS} &
\textbf{GPT-4o FS} &
\textbf{Claude 4.5 ZS} &
\textbf{Claude 4.5 FS} \\
\midrule
F1 Score (\%) & $94.42\pm0.10$ & $93.49\pm0.11$ & $93.59\pm0.23$ & $87.00\pm0.07$ & $90.45\pm0.03$ & $91.78\pm0.01$ & $94.41\pm0.06$ \\
Precision (\%) & $94.40\pm0.08$ & $93.46\pm0.09$ & $93.65\pm0.25$ & $88.49\pm0.06$ & $90.94\pm0.02$ & $91.99\pm0.01$ & $94.28\pm0.06$ \\
Recall (\%) & $94.44\pm0.13$ & $93.53\pm0.14$ & $93.54\pm0.22$ & $88.29\pm0.06$ & $91.35\pm0.03$ & $92.53\pm0.01$ & $94.79\pm0.05$ \\
Accuracy (\%) & $94.49\pm0.10$ & $93.58\pm0.10$ & $93.69\pm0.23$ & $87.00\pm0.07$ & $90.46\pm0.03$ & $91.80\pm0.01$ & $94.45\pm0.06$ \\
\midrule
Inference Latency p50 (ms) & 147.03 & 97.88 & 133.38 & 377.10 & 326.00 & 1394.27 & 1109.11 \\
Inference Latency p95 (ms) & 192.38 & 123.93 & 207.02 & 591.46 & 518.82 & 2047.78 & 1895.08 \\
Inference Latency p99 (ms) & 226.67 & 142.65 & 249.28 & 1056.93 & 1094.62 & 2755.15 & 2456.68 \\
\midrule
TTFT p50 (ms) & N/A & N/A & N/A & 367.01 & 323.21 & 1328.16 & 1045.73 \\
TTFT p95 (ms) & N/A & N/A & N/A & 581.83 & 515.90 & 1944.12 & 1775.75 \\
TTFT p99 (ms) & N/A & N/A & N/A & 1040.02 & 1091.55 & 2590.21 & 2308.00 \\
\midrule
Avg. Input Tokens / Req & N/A & N/A & N/A & 72.99 & 150.99 & 83.89 & 174.89 \\
Avg. Output Tokens / Req & N/A & N/A & N/A & 1.00 & 1.00 & 5.00 & 5.00 \\
Warm-up Iterations & 10 & 10 & 10 & 10 & 10 & 10 & 10 \\
Sampling Temperature & N/A & N/A & N/A & 0.00 & 0.00 & 0.00 & 0.00 \\
Top-p Sampling & N/A & N/A & N/A & 1.00 & 1.00 & N/A & N/A \\
\midrule
Training / Inference Setup & FT & FT & FT & ZS & FS & ZS & FS \\
\midrule
Input Token Price (USD / 1M) & N/A & N/A & N/A & 2.50 & 2.50 & 3.00 & 3.00 \\
Output Token Price (USD / 1M) & N/A & N/A & N/A & 10.00 & 10.00 & 15.00 & 15.00 \\
Allocated CPU (vCPU) & 2 & 2 & 2 & N/A & N/A & N/A & N/A \\
Allocated Memory (GiB) & 2 & 2 & 2 & N/A & N/A & N/A & N/A \\
CPU Price (USD / 1M vCPU-s) & 24.00 & 24.00 & 24.00 & N/A & N/A & N/A & N/A \\
Memory Price (USD / 1M GiB-s) & 2.50 & 2.50 & 2.50 & N/A & N/A & N/A & N/A \\
\midrule
Estimated Cost (USD / 1M req) & 7.79 & 5.19 & 7.07 & 192.48 & 387.48 & 326.67 & 599.67 \\
\bottomrule
\end{tabular}}
\end{table*}

\begin{table*}[t]
\caption{AG News — Topic classification results (mean $\pm$ std over three independent repetitions; three seeds for encoders, three API runs for LLM prompting).}
\label{tab:agnews}
\centering
\resizebox{\textwidth}{!}{
\begin{tabular}{lccccccc}
\toprule
\textbf{Metric} &
\textbf{BERT} &
\textbf{DistilBERT} &
\textbf{RoBERTa} &
\textbf{GPT-4o ZS} &
\textbf{GPT-4o FS} &
\textbf{Claude 4.5 ZS} &
\textbf{Claude 4.5 FS} \\
\midrule
F1 Score (\%) & $94.43\pm0.10$ & $94.11\pm0.07$ & $94.63\pm0.14$ & $87.93\pm0.25$ & $89.65\pm0.13$ & $91.35\pm0.11$ & $90.56\pm0.06$ \\
Precision (\%) & $94.46\pm0.09$ & $94.14\pm0.09$ & $94.65\pm0.13$ & $88.22\pm0.24$ & $89.73\pm0.14$ & $91.40\pm0.11$ & $90.60\pm0.06$ \\
Recall (\%) & $94.42\pm0.09$ & $94.11\pm0.07$ & $94.62\pm0.14$ & $88.00\pm0.25$ & $89.68\pm0.13$ & $91.36\pm0.11$ & $90.59\pm0.06$ \\
Accuracy (\%) & $94.42\pm0.09$ & $94.11\pm0.09$ & $94.62\pm0.14$ & $88.00\pm0.25$ & $89.68\pm0.13$ & $91.36\pm0.11$ & $90.59\pm0.06$ \\
\midrule
Inference Latency p50 (ms) & 196.91 & 108.19 & 188.70 & 410.81 & 332.31 & 1434.82 & 1002.56 \\
Inference Latency p95 (ms) & 285.80 & 161.32 & 279.20 & 693.55 & 582.24 & 2298.01 & 2544.03 \\
Inference Latency p99 (ms) & 572.99 & 515.73 & 485.33 & 1249.39 & 955.58 & 4551.21 & 5756.15 \\
\midrule
TTFT p50 (ms) & N/A & N/A & N/A & 404.67 & 328.70 & 1363.53 & 916.85 \\
TTFT p95 (ms) & N/A & N/A & N/A & 686.61 & 578.81 & 2212.85 & 2441.21 \\
TTFT p99 (ms) & N/A & N/A & N/A & 1241.98 & 954.27 & 4477.20 & 5719.61 \\
\midrule
Avg. Input Tokens / Req & N/A & N/A & N/A & 106.40 & 357.40 & 121.86 & 398.86 \\
Avg. Output Tokens / Req & N/A & N/A & N/A & 1.00 & 1.00 & 5.00 & 5.00 \\
Warm-up Iterations & 10 & 10 & 10 & 10 & 10 & 10 & 10 \\
Sampling Temperature & N/A & N/A & N/A & 0.00 & 0.00 & 0.00 & 0.00 \\
Top-p Sampling & N/A & N/A & N/A & 1.00 & 1.00 & N/A & N/A \\
\midrule
Training / Inference Setup & FT & FT & FT & ZS & FS & ZS & FS \\
\midrule
Input Token Price (USD / 1M) & N/A & N/A & N/A & 2.50 & 2.50 & 3.00 & 3.00 \\
Output Token Price (USD / 1M) & N/A & N/A & N/A & 10.00 & 10.00 & 15.00 & 15.00 \\
Allocated CPU (vCPU) & 2 & 2 & 2 & N/A & N/A & N/A & N/A \\
Allocated Memory (GiB) & 2 & 2 & 2 & N/A & N/A & N/A & N/A \\
CPU Price (USD / 1M vCPU-s) & 24.00 & 24.00 & 24.00 & N/A & N/A & N/A & N/A \\
Memory Price (USD / 1M GiB-s) & 2.50 & 2.50 & 2.50 & N/A & N/A & N/A & N/A \\
\midrule
Estimated Cost (USD / 1M req) & 10.44 & 5.73 & 10.00 & 276.00 & 903.50 & 440.58 & 1271.58 \\
\bottomrule
\end{tabular}}
\end{table*}

\begin{table*}[t]
\caption{DBPedia — Ontology classification results (mean $\pm$ std over three independent repetitions; three seeds for encoders, three API runs for LLM prompting).}
\label{tab:dbpedia}
\centering
\resizebox{\textwidth}{!}{
\begin{tabular}{lccccccc}
\toprule
\textbf{Metric} &
\textbf{BERT} &
\textbf{DistilBERT} &
\textbf{RoBERTa} &
\textbf{GPT-4o ZS} &
\textbf{GPT-4o FS} &
\textbf{Claude 4.5 ZS} &
\textbf{Claude 4.5 FS} \\
\midrule
F1 Score (\%) & $99.40\pm0.04$ & $99.40\pm0.01$ & $99.33\pm0.06$ & $96.12\pm0.03$ & $97.06\pm0.07$ & $98.83\pm0.04$ & $98.39\pm0.04$ \\
Precision (\%) & $99.40\pm0.04$ & $99.40\pm0.01$ & $99.33\pm0.06$ & $96.33\pm0.03$ & $97.17\pm0.06$ & $98.84\pm0.04$ & $98.41\pm0.04$ \\
Recall (\%) & $99.40\pm0.04$ & $99.40\pm0.01$ & $99.33\pm0.06$ & $96.14\pm0.03$ & $97.05\pm0.07$ & $98.83\pm0.04$ & $98.39\pm0.04$ \\
Accuracy (\%) & $99.40\pm0.04$ & $99.40\pm0.01$ & $99.33\pm0.06$ & $96.14\pm0.03$ & $97.05\pm0.07$ & $98.83\pm0.04$ & $98.39\pm0.04$ \\
\midrule
Inference Latency p50 (ms) & 203.23 & 132.57 & 205.63 & 406.32 & 417.39 & 1127.83 & 1126.59 \\
Inference Latency p95 (ms) & 361.28 & 217.08 & 381.12 & 610.27 & 653.21 & 1676.71 & 1862.52 \\
Inference Latency p99 (ms) & 782.80 & 306.74 & 673.28 & 926.50 & 985.09 & 2394.98 & 2400.27 \\
\midrule
TTFT p50 (ms) & N/A & N/A & N/A & 403.77 & 415.08 & 1058.00 & 1044.10 \\
TTFT p95 (ms) & N/A & N/A & N/A & 609.24 & 650.90 & 1600.58 & 1775.36 \\
TTFT p99 (ms) & N/A & N/A & N/A & 922.85 & 984.43 & 2318.16 & 2295.27 \\
\midrule
Avg. Input Tokens / Req & N/A & N/A & N/A & 181.28 & 757.28 & 225.63 & 875.63 \\
Avg. Output Tokens / Req & N/A & N/A & N/A & 1.00 & 1.00 & 5.00 & 5.00 \\
Warm-up Iterations & 10 & 10 & 10 & 10 & 10 & 10 & 10 \\
Sampling Temperature & N/A & N/A & N/A & 0.00 & 0.00 & 0.00 & 0.00 \\
Top-p Sampling & N/A & N/A & N/A & 1.00 & 1.00 & N/A & N/A \\
\midrule
Training / Inference Setup & FT & FT & FT & ZS & FS & ZS & FS \\
\midrule
Input Token Price (USD / 1M) & N/A & N/A & N/A & 2.50 & 2.50 & 3.00 & 3.00 \\
Output Token Price (USD / 1M) & N/A & N/A & N/A & 10.00 & 10.00 & 15.00 & 15.00 \\
Allocated CPU (vCPU) & 2 & 2 & 2 & N/A & N/A & N/A & N/A \\
Allocated Memory (GiB) & 2 & 2 & 2 & N/A & N/A & N/A & N/A \\
CPU Price (USD / 1M vCPU-s) & 24.00 & 24.00 & 24.00 & N/A & N/A & N/A & N/A \\
Memory Price (USD / 1M GiB-s) & 2.50 & 2.50 & 2.50 & N/A & N/A & N/A & N/A \\
\midrule
Estimated Cost (USD / 1M req) & 10.77 & 7.03 & 10.90 & 463.20 & 1903.20 & 751.89 & 2701.89 \\
\bottomrule
\end{tabular}}
\end{table*}

\clearpage

\subsection{Rationale Behind the Utility Function}
\label{sec:utility_rationale}

The utility function introduced in this work is not intended as a universal objective, but as a pragmatic decision rule designed to reflect common operational priorities in production NLP systems. Its form is guided by three desiderata: monotonicity with respect to predictive performance, explicit penalization of recurring inference cost, and a smooth, interpretable treatment of latency constraints.

We define utility as
\[
U(F1, Cost, Latency) = \frac{F1}{Cost} \cdot \exp\!\left(-\frac{Latency_{50}}{\tau}\right),
\]
where $F1$ captures predictive reliability, $Cost$ denotes per-request monetary cost under steady-state inference, $Latency_{50}$ is the median end-to-end response time, and $\tau$ represents an application-specific latency tolerance.

\paragraph{Accuracy--cost trade-off.}
The ratio $F1 / Cost$ directly reflects the notion of \emph{predictive return on investment}: improvements in classification quality must justify their associated recurring inference cost. Unlike additive formulations, this ratio naturally emphasizes efficiency at scale, where small per-request cost differences compound into substantial budgetary impact under high-throughput workloads. Models that achieve marginal accuracy gains at disproportionate cost are therefore explicitly penalized.

\paragraph{Latency as a soft constraint.}
Latency enters the utility through an exponential decay term rather than a hard threshold or linear penalty. This choice reflects the operational reality of production systems, where latency violations do not occur at a single cutoff, but instead degrade user experience and system reliability progressively. The exponential form ensures that latency penalties increase smoothly and monotonically, while the tolerance parameter $\tau$ enables sensitivity analysis across deployment regimes ranging from interactive to batch-oriented inference.

\paragraph{Scale invariance and interpretability.}
By separating accuracy--cost efficiency from latency sensitivity, the utility function avoids arbitrary weighting between incommensurate quantities. The formulation remains invariant to linear rescaling of cost units and yields values that are straightforward to interpret comparatively, even if the absolute magnitude of utility is not intrinsically meaningful. This property is particularly useful for ranking models across heterogeneous deployment contexts.

\paragraph{Alternatives and limitations.}
We deliberately avoid weighted linear combinations of accuracy, latency, and cost, as such formulations require arbitrary coefficient selection and often obscure trade-offs through compensatory effects. Multi-objective optimization techniques, including Pareto dominance, are therefore used in parallel to characterize non-dominated regimes without collapsing objectives. Utility-based ranking is employed only when a single deployment decision must be made under fixed operational constraints.

Overall, this utility function serves as a transparent and defensible decision heuristic rather than a normative optimization target. Its purpose is to make trade-offs explicit and comparable, complementing the Pareto-based analysis and supporting principled model selection in cost- and latency-constrained production environments.

\subsection{Utility-based Model Selection}
\label{sec:utility_selection}

To translate empirical results into concrete deployment decisions, we rank models using the utility function introduced in the Method section:

\[
U(F1, Cost, Latency) = \frac{F1}{Cost} \cdot \exp\left(-\frac{Latency_{50}}{\tau}\right),
\]
where $Latency_{50}$ denotes p50 end-to-end inference latency and $\tau$ represents a latency tolerance parameter controlling the relative importance of responsiveness. We evaluate this utility under multiple operational regimes, with $\tau \in \{250, 500, 1000\}$ ms, spanning latency-sensitive (interactive) to latency-tolerant (batch or asynchronous) deployments.

This formulation yields a scalar ranking that makes trade-offs explicit: improvements in predictive performance must justify their associated increases in latency and cost. By varying $\tau$, we assess the robustness of model preferences across deployment contexts, rather than committing to a single latency assumption. As a result, utility-based ranking provides a concrete decision rule for practitioners who must select a single model under fixed infrastructure and budget constraints.

Tables~\ref{tab:utility_imdb}--\ref{tab:utility_dbpedia} report utility-based rankings for each dataset under varying latency tolerances, with values corresponding to $100\times U$ for readability; higher values indicate more favorable accuracy--latency--cost trade-offs.

\begin{table}[h]
\caption{IMDB — Utility-based model ranking under different latency tolerances $\tau$. 
Reported values correspond to $100\times U$ for readability; higher utility is better. 
Rank shown in parentheses.}
\label{tab:utility_imdb}
\centering
\begin{tabular}{lccc}
\toprule
\textbf{Model} &
$\boldsymbol{\tau=250\,ms}$ &
$\boldsymbol{\tau=500\,ms}$ &
$\boldsymbol{\tau=1000\,ms}$ \\
\midrule
BERT            & 0.54 (2) & 1.41 (2) & 2.27 (2) \\
DistilBERT      & 2.91 (1) & 4.66 (1) & 5.89 (1) \\
RoBERTa         & 0.24 (3) & 0.83 (3) & 1.54 (3) \\
GPT-4o (ZS)     & 0.03 (4) & 0.06 (4) & 0.08 (4) \\
GPT-4o (FS)     & 0.01 (5) & 0.03 (5) & 0.04 (5) \\
Claude 4.5 (ZS) & 0.00 (6) & 0.01 (6) & 0.02 (6) \\
Claude 4.5 (FS) & 0.00 (7) & 0.00 (7) & 0.01 (7) \\
\bottomrule
\end{tabular}
\end{table}

\begin{table}[h]
\caption{SST-2 — Utility-based model ranking under different latency tolerances $\tau$. 
Reported values correspond to $100\times U$ for readability; higher utility is better. 
Rank shown in parentheses.}
\label{tab:utility_sst2}
\centering
\begin{tabular}{lccc}
\toprule
\textbf{Model} &
$\boldsymbol{\tau=250\,ms}$ &
$\boldsymbol{\tau=500\,ms}$ &
$\boldsymbol{\tau=1000\,ms}$ \\
\midrule
BERT            & 6.73 (3) & 9.03 (3) & 10.46 (3) \\
DistilBERT      & 12.18 (1) & 14.81 (1) & 16.33 (1) \\
RoBERTa         & 7.76 (2) & 10.14 (2) & 11.58 (2) \\
GPT-4o (ZS)     & 0.10 (4) & 0.21 (4) & 0.31 (4) \\
GPT-4o (FS)     & 0.06 (5) & 0.12 (5) & 0.17 (5) \\
Claude 4.5 (ZS) & 0.00 (7) & 0.02 (6) & 0.07 (6) \\
Claude 4.5 (FS) & 0.00 (6) & 0.02 (7) & 0.05 (7) \\
\bottomrule
\end{tabular}
\end{table}

\begin{table}[h]
\caption{AG News — Utility-based model ranking under different latency tolerances $\tau$. 
Reported values correspond to $100\times U$ for readability; higher utility is better. 
Rank shown in parentheses.}
\label{tab:utility_agnews}
\centering
\begin{tabular}{lccc}
\toprule
\textbf{Model} &
$\boldsymbol{\tau=250\,ms}$ &
$\boldsymbol{\tau=500\,ms}$ &
$\boldsymbol{\tau=1000\,ms}$ \\
\midrule
BERT            & 4.11 (3) & 6.10 (3) & 7.43 (3) \\
DistilBERT      & 10.65 (1) & 13.23 (1) & 14.74 (1) \\
RoBERTa         & 4.45 (2) & 6.49 (2) & 7.84 (2) \\
GPT-4o (ZS)     & 0.06 (4) & 0.14 (4) & 0.21 (4) \\
GPT-4o (FS)     & 0.03 (5) & 0.05 (5) & 0.07 (5) \\
Claude 4.5 (ZS) & 0.00 (7) & 0.01 (6) & 0.05 (6) \\
Claude 4.5 (FS) & 0.00 (6) & 0.01 (7) & 0.03 (7) \\
\bottomrule
\end{tabular}
\end{table}

\begin{table}[h]
\caption{DBPedia — Utility-based model ranking under different latency tolerances $\tau$. 
Reported values correspond to $100\times U$ for readability; higher utility is better. 
Rank shown in parentheses.}
\label{tab:utility_dbpedia}
\centering
\begin{tabular}{lccc}
\toprule
\textbf{Model} &
$\boldsymbol{\tau=250\,ms}$ &
$\boldsymbol{\tau=500\,ms}$ &
$\boldsymbol{\tau=1000\,ms}$ \\
\midrule
BERT            & 4.09 (2) & 6.15 (2) & 7.53 (2) \\
DistilBERT      & 8.32 (1) & 10.85 (1) & 12.38 (1) \\
RoBERTa         & 4.00 (3) & 6.04 (3) & 7.42 (3) \\
GPT-4o (ZS)     & 0.04 (4) & 0.09 (4) & 0.14 (4) \\
GPT-4o (FS)     & 0.01 (5) & 0.02 (5) & 0.03 (6) \\
Claude 4.5 (ZS) & 0.00 (6) & 0.01 (6) & 0.04 (5) \\
Claude 4.5 (FS) & 0.00 (7) & 0.00 (7) & 0.01 (7) \\
\bottomrule
\end{tabular}
\end{table}

\paragraph{Summary and implications.}
Across all datasets and latency regimes, utility-based ranking yields a remarkably stable and interpretable ordering.  Fine-tuned encoders consistently dominate LLM-based prompting by one to two orders of magnitude in utility ($100\times U$), even under relaxed latency tolerances ($\tau=1000$\,ms), indicating that these differences are not merely rank-level but quantitatively substantial.  In particular, DistilBERT emerges as the top-ranked model in every dataset and for all values of $\tau$, reflecting a favorable balance between near-ceiling accuracy, low inference latency, and minimal operating cost. While RoBERTa occasionally attains higher raw F1, its increased p50 latency reduces utility under stricter latency budgets, illustrating how marginal accuracy gains can be outweighed once responsiveness is treated as a first-class constraint.

In contrast, LLM prompting—especially in few-shot configurations—achieves substantially lower utility across all regimes. Although larger $\tau$ values partially attenuate latency penalties, high per-request cost remains the dominant limiting factor, preventing LLM-based approaches from becoming competitive under realistic deployment constraints. Taken together, these results show that utility-based selection provides a robust, deployment-aware decision criterion: it favors compact, specialized models for fixed-label classification tasks, even when raw accuracy differences appear small in isolation but translate into large disparities in end-to-end system efficiency.

\subsection{Pareto-based Model Selection}
\label{sec:pareto_selection}

Complementary to scalar utility ranking, we analyze model selection through Pareto dominance in the $(F1, Latency, Cost)$ space. A configuration is considered Pareto-optimal if no other model simultaneously achieves higher predictive performance while incurring lower inference latency and lower cost. Unlike utility-based ranking, Pareto analysis is regime-independent and does not require specifying an explicit latency tolerance or cost weighting.

To facilitate a systematic comparison across tasks, we present all Pareto projections jointly. For each dataset, we report three two-dimensional views: macro-F1 versus estimated inference cost, estimated cost versus inference latency, and macro-F1 versus latency. All twelve plots are grouped together for visual consistency and to enable cross-dataset comparison under a unified scale.

\begin{figure*}[t]
    \centering
    \begin{subfigure}[t]{\subfigwidth}
        \includegraphics[width=\textwidth]{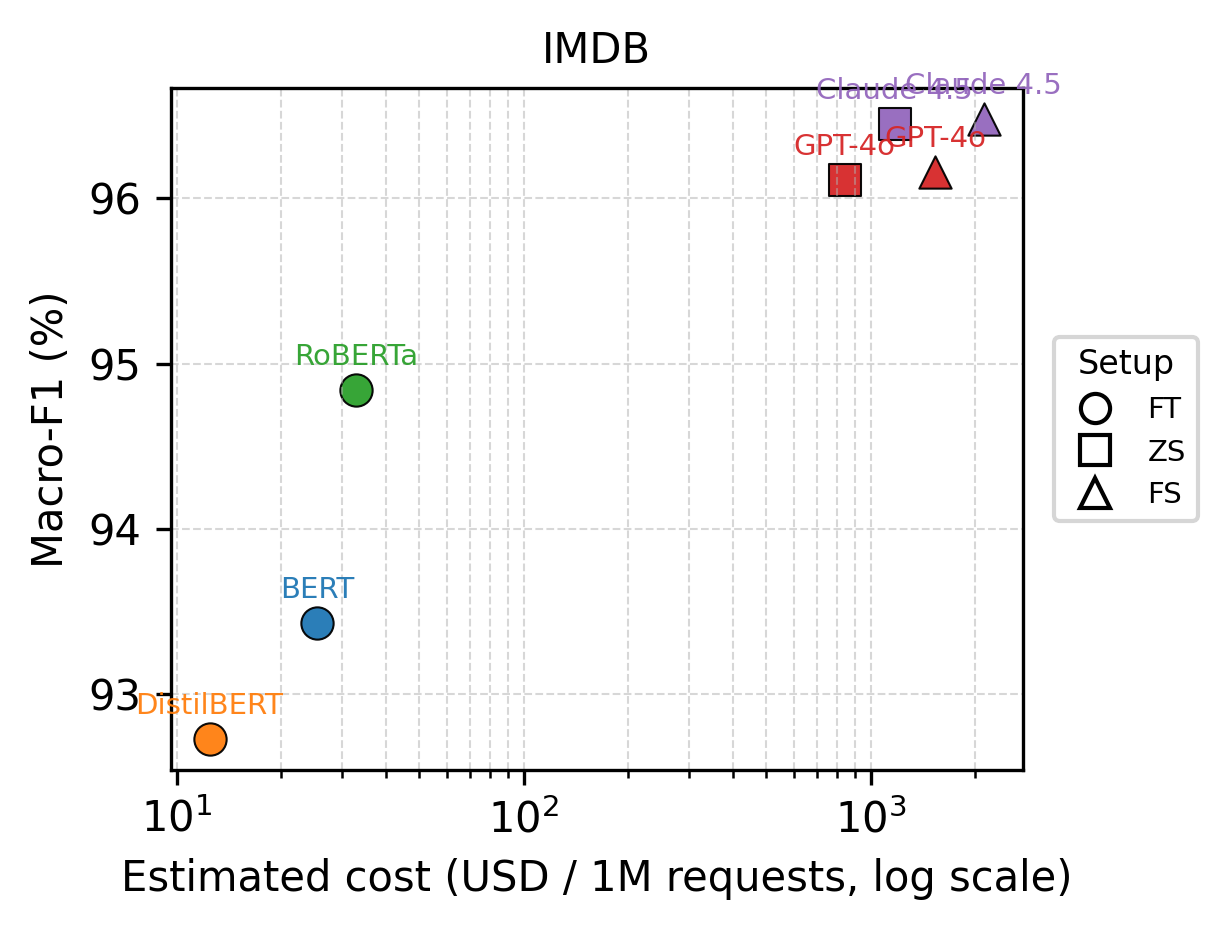}
        \caption{F1 vs. cost}
        \label{fig:imdb_f1_cost}
    \end{subfigure}
    \hfill
    \begin{subfigure}[t]{\subfigwidth}
        \includegraphics[width=\textwidth]{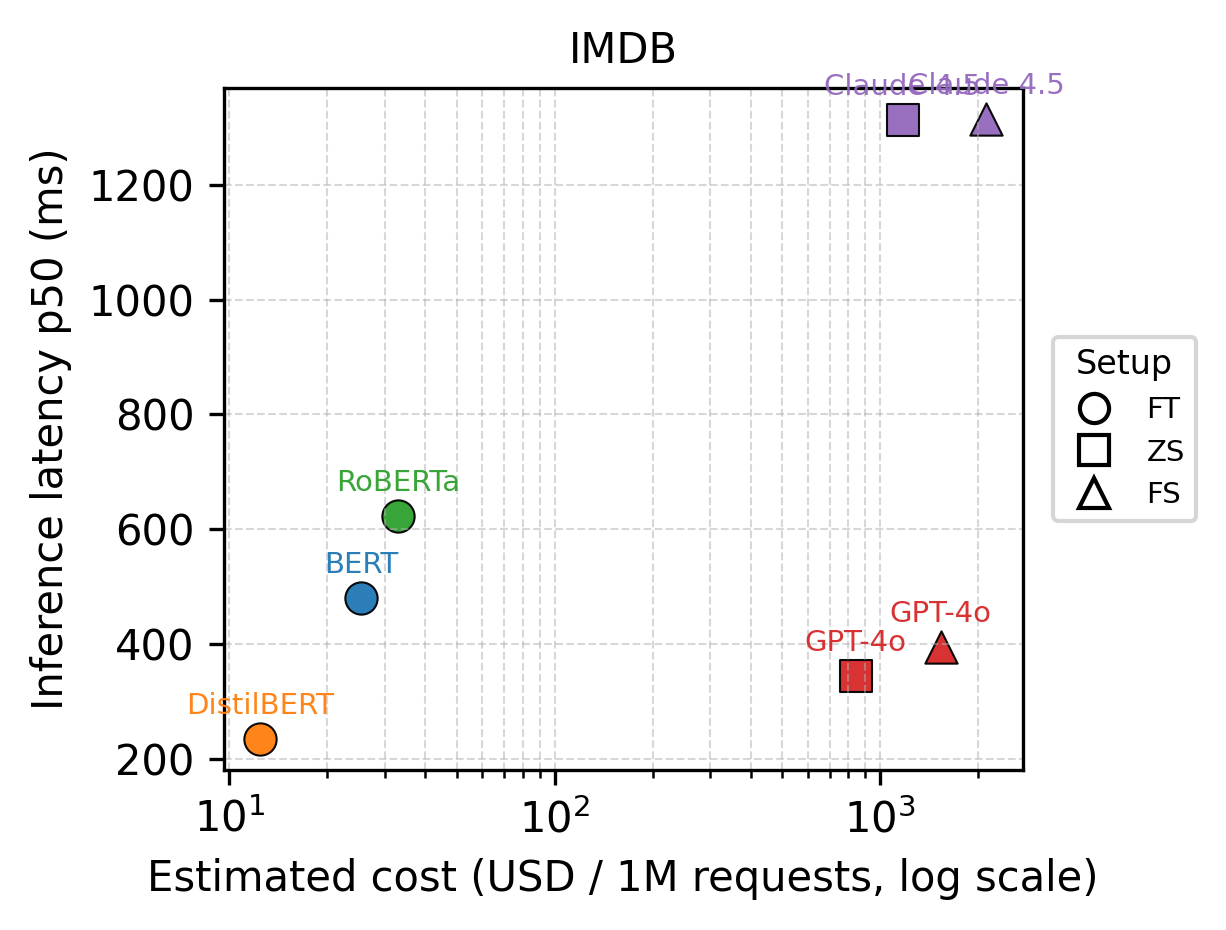}
        \caption{Cost vs. latency}
        \label{fig:imdb_cost_latency}
    \end{subfigure}
    \hfill
    \begin{subfigure}[t]{\subfigwidth}
        \includegraphics[width=\textwidth]{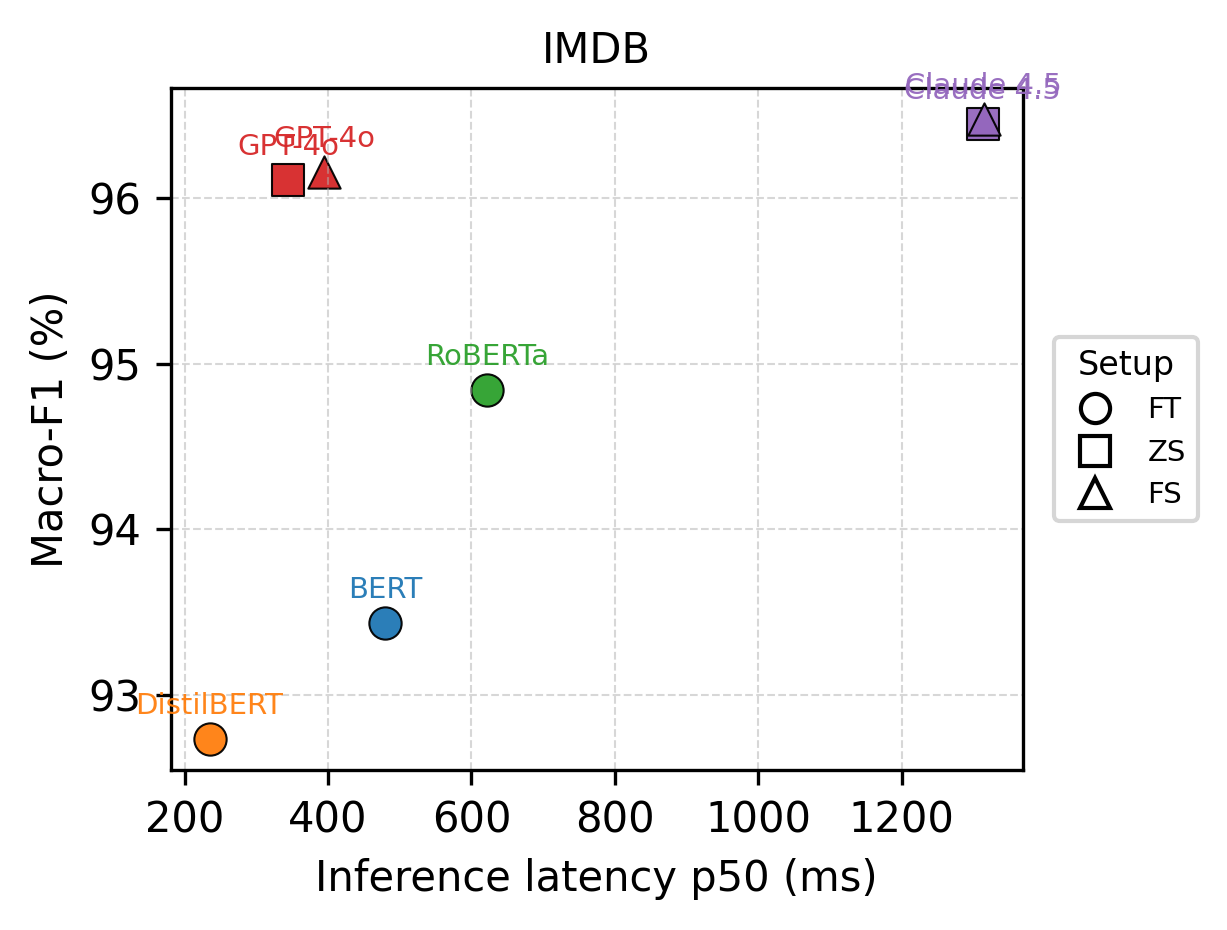}
        \caption{F1 vs. latency}
        \label{fig:imdb_f1_latency}
    \end{subfigure}
    \caption{Pareto projections for IMDB sentiment classification. Each point represents a model configuration; colors denote model families and marker shapes indicate training or prompting setup.}
    \label{fig:pareto_imdb}
\end{figure*}

\begin{figure*}[t]
    \centering
    \begin{subfigure}[t]{\subfigwidth}
        \includegraphics[width=\textwidth]{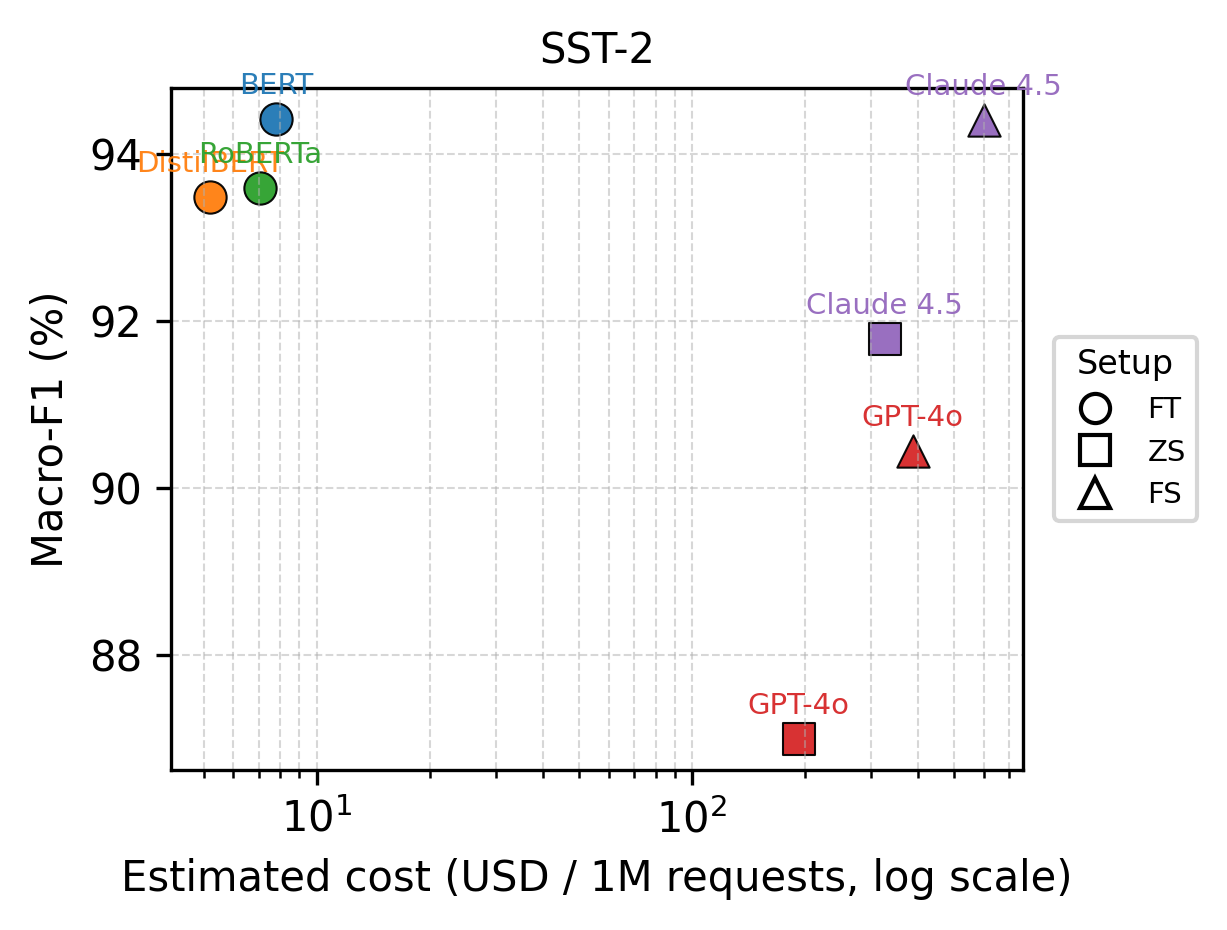}
        \caption{F1 vs. cost}
        \label{fig:sst2_f1_cost}
    \end{subfigure}
    \hfill
    \begin{subfigure}[t]{\subfigwidth}
        \includegraphics[width=\textwidth]{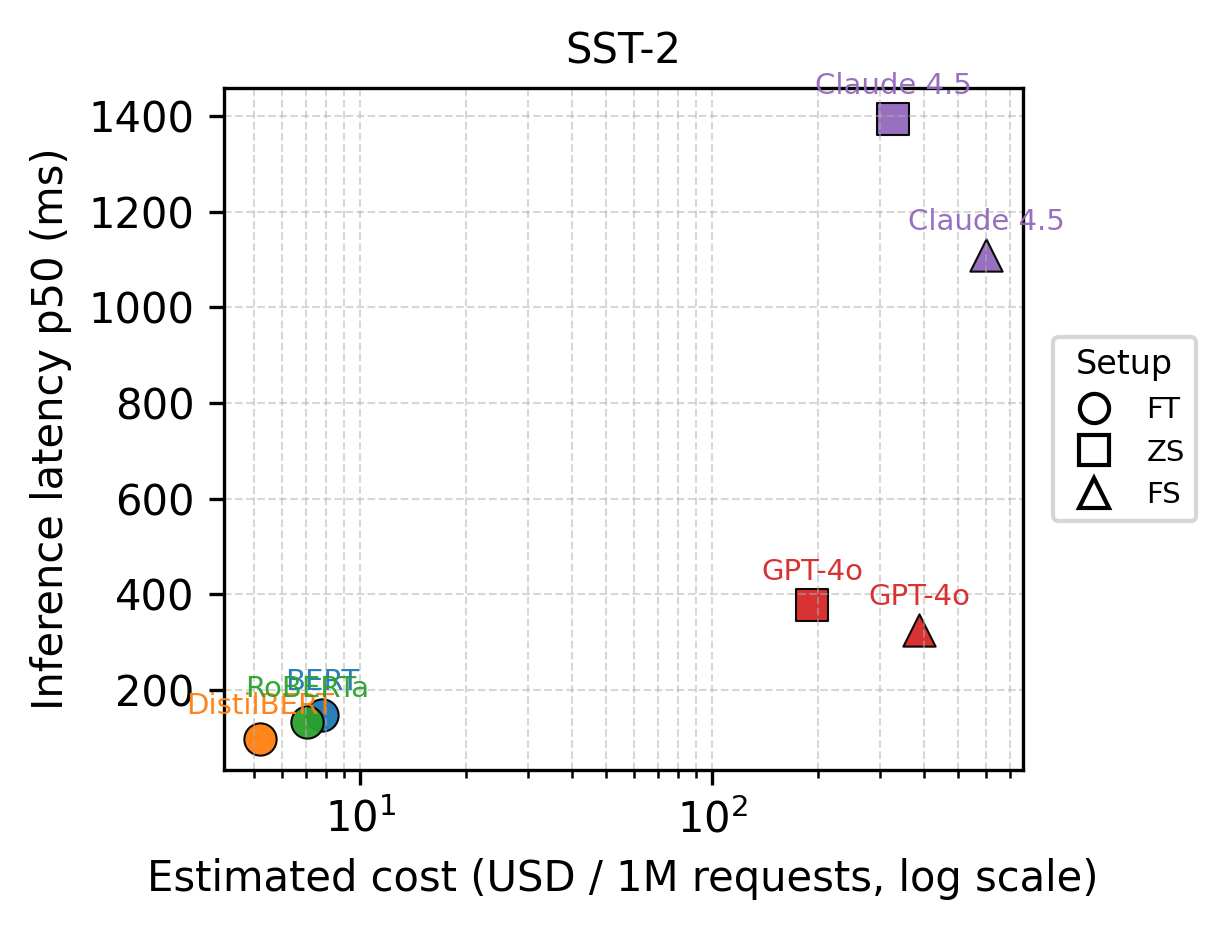}
        \caption{Cost vs. latency}
        \label{fig:sst2_cost_latency}
    \end{subfigure}
    \hfill
    \begin{subfigure}[t]{\subfigwidth}
        \includegraphics[width=\textwidth]{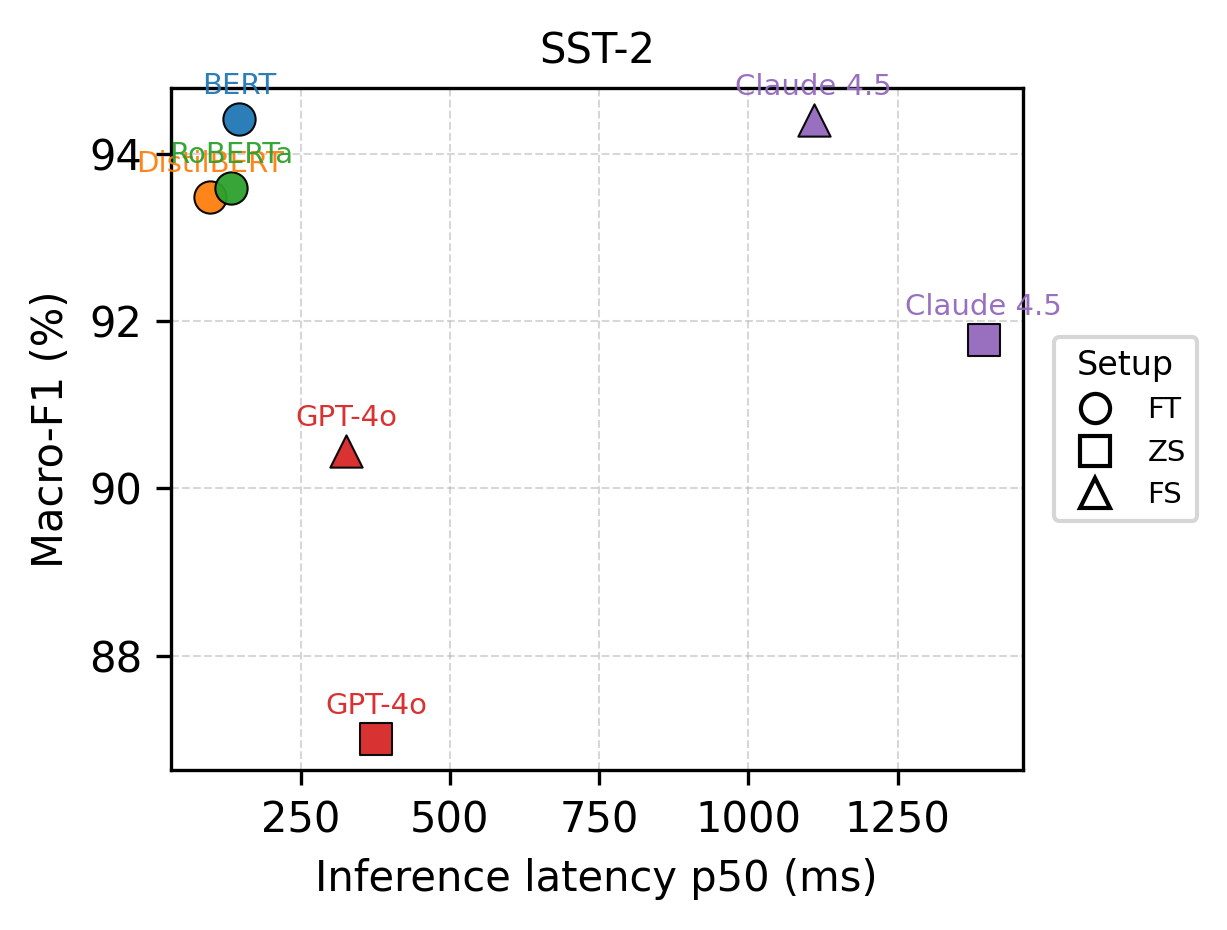}
        \caption{F1 vs. latency}
        \label{fig:sst2_f1_latency}
    \end{subfigure}
    \caption{Pareto projections for SST-2 sentiment classification under cost and latency constraints.}
    \label{fig:pareto_sst2}
\end{figure*}

\begin{figure*}[t]
    \centering
    \begin{subfigure}[t]{\subfigwidth}
        \includegraphics[width=\textwidth]{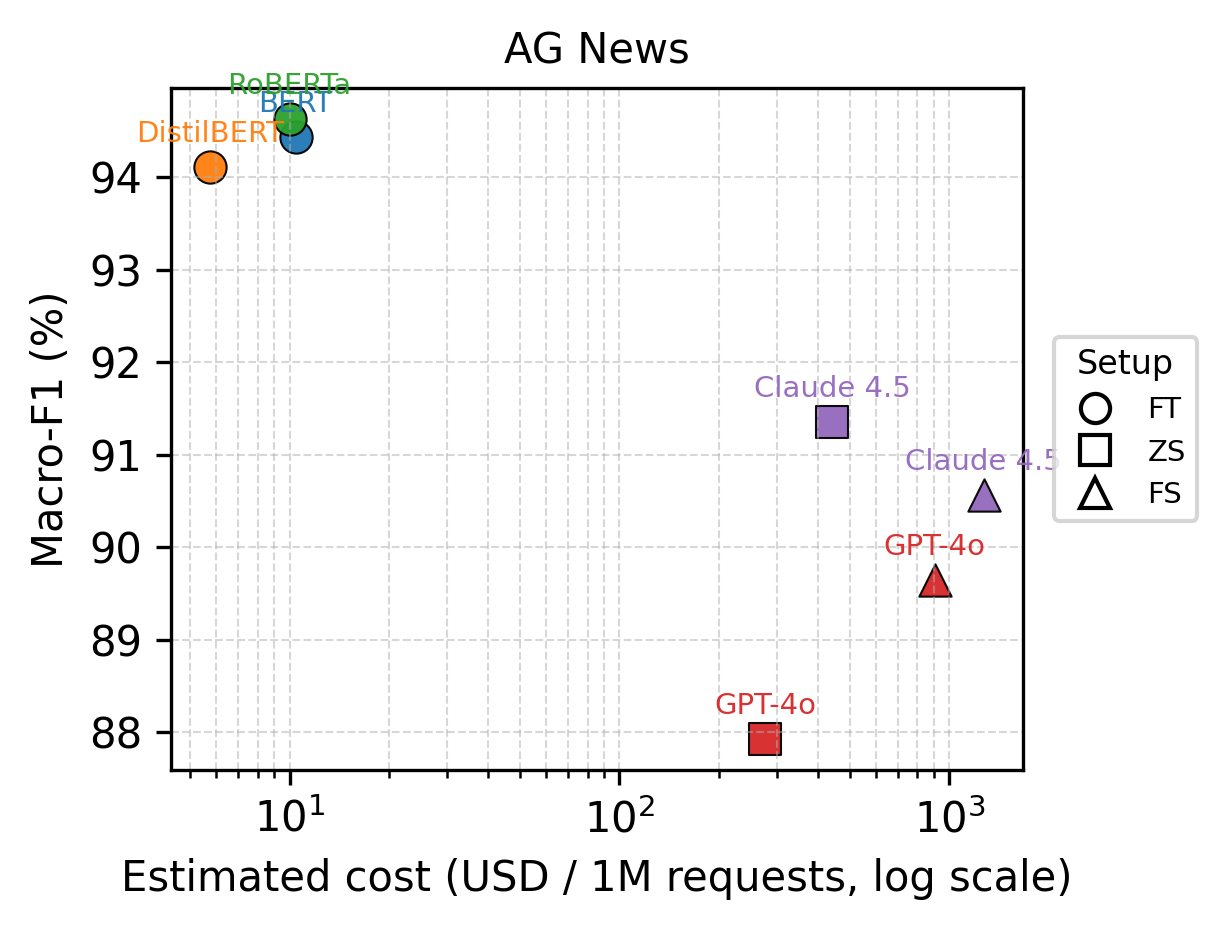}
        \caption{F1 vs. cost}
        \label{fig:agnews_f1_cost}
    \end{subfigure}
    \hfill
    \begin{subfigure}[t]{\subfigwidth}
        \includegraphics[width=\textwidth]{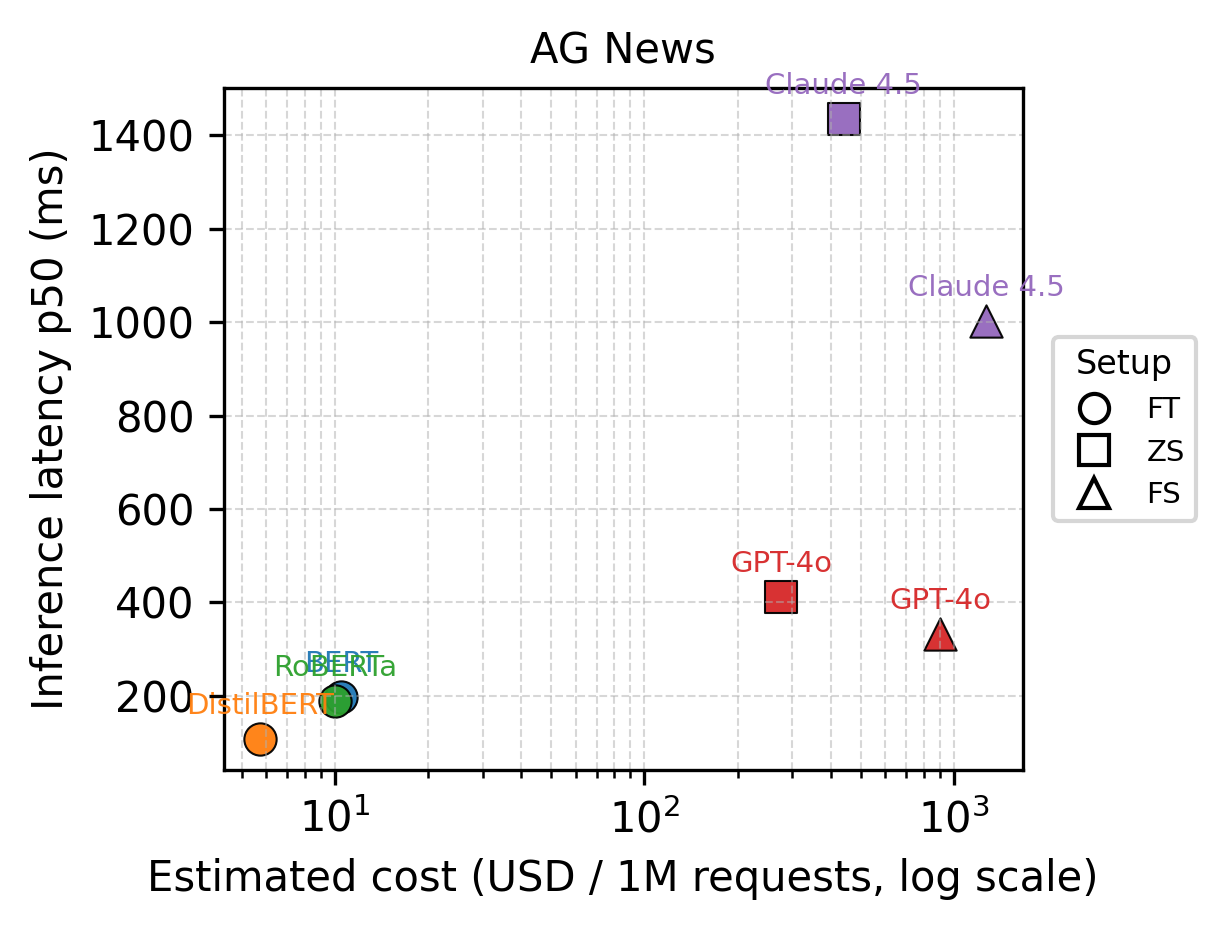}
        \caption{Cost vs. latency}
        \label{fig:agnews_cost_latency}
    \end{subfigure}
    \hfill
    \begin{subfigure}[t]{\subfigwidth}
        \includegraphics[width=\textwidth]{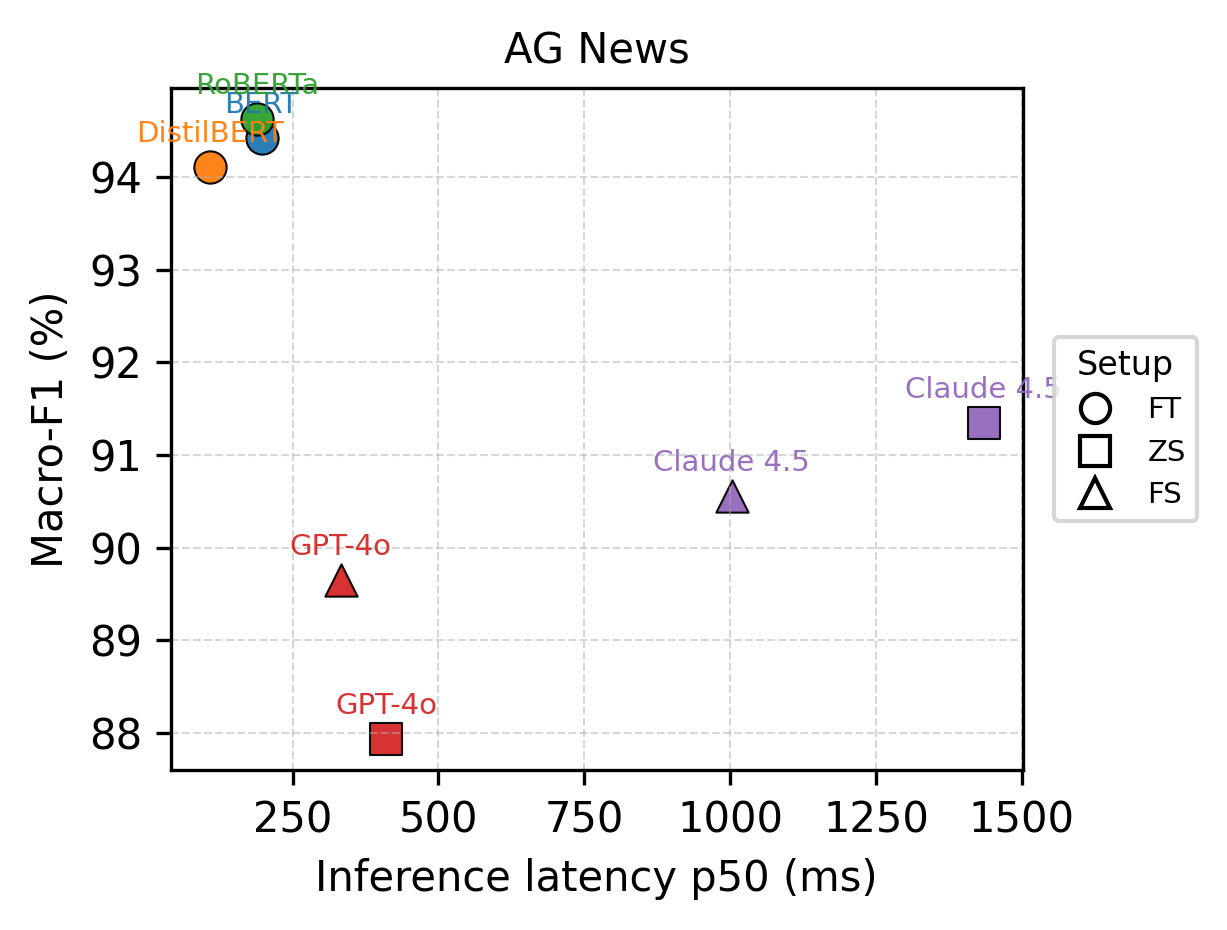}
        \caption{F1 vs. latency}
        \label{fig:agnews_f1_latency}
    \end{subfigure}
    \caption{Pareto projections for AG News topic classification, illustrating accuracy–efficiency trade-offs in multi-class settings.}
    \label{fig:pareto_agnews}
\end{figure*}

\begin{figure*}[t]
    \centering
    \begin{subfigure}[t]{\subfigwidth}
        \includegraphics[width=\textwidth]{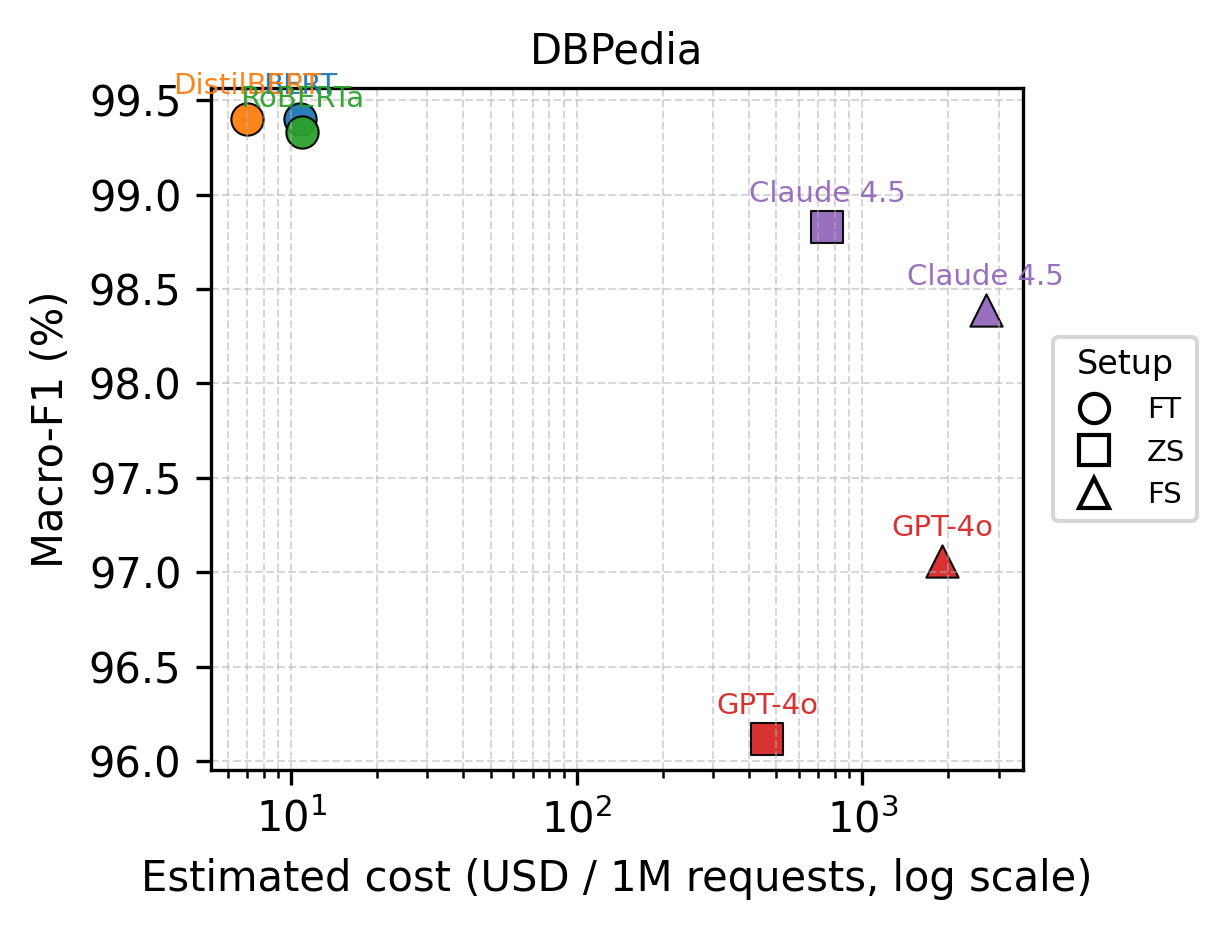}
        \caption{F1 vs. cost}
        \label{fig:dbpedia_f1_cost}
    \end{subfigure}
    \hfill
    \begin{subfigure}[t]{\subfigwidth}
        \includegraphics[width=\textwidth]{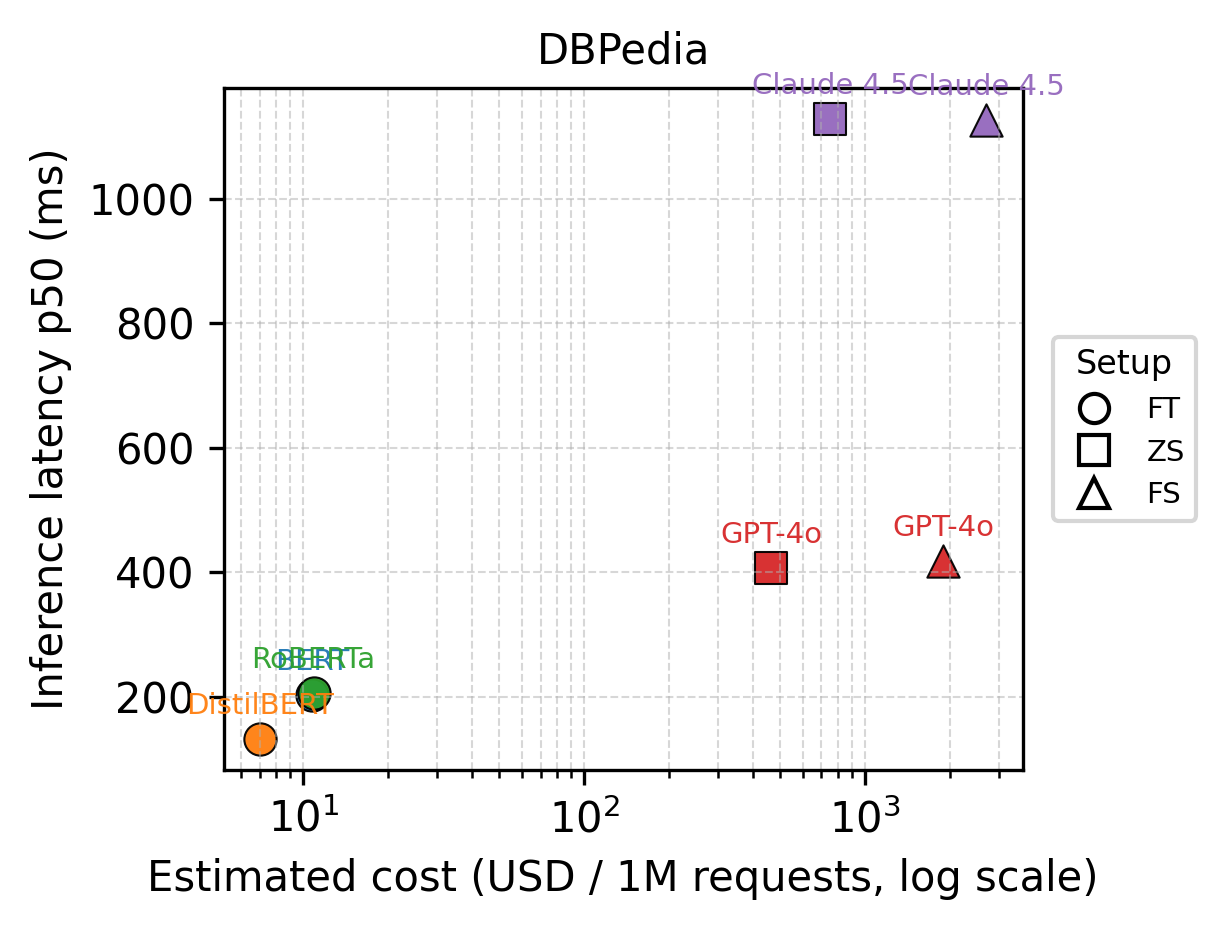}
        \caption{Cost vs. latency}
        \label{fig:dbpedia_cost_latency}
    \end{subfigure}
    \hfill
    \begin{subfigure}[t]{\subfigwidth}
        \includegraphics[width=\textwidth]{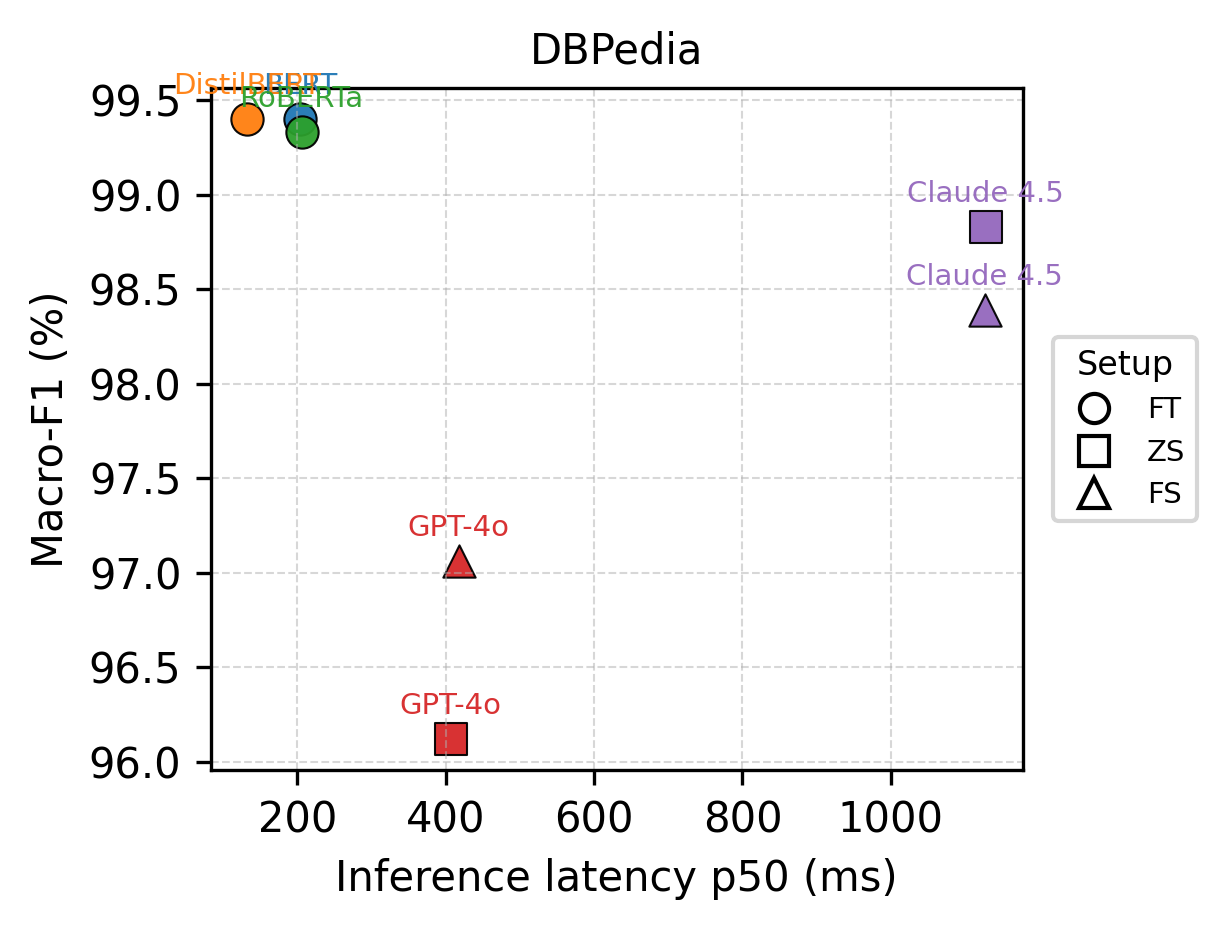}
        \caption{F1 vs. latency}
        \label{fig:dbpedia_f1_latency}
    \end{subfigure}
    \caption{Pareto projections for DBPedia ontology classification. Encoder-based models saturate performance while remaining dominant in cost and latency.}
    \label{fig:pareto_dbpedia}
\end{figure*}

\paragraph{Pareto frontier projections.}
Figures~\ref{fig:pareto_imdb}--\ref{fig:pareto_dbpedia} jointly visualize Pareto trade-offs across all four datasets. Across benchmarks, fine-tuned encoder models consistently occupy a low-cost, low-latency region while achieving strong macro-F1 performance. In contrast, large language model (LLM) prompting configurations populate a substantially higher-cost regime, with latency increasing further for models relying on few-shot prompting.

The cost--latency projections reveal a clear separation between encoder-based inference, which remains bounded below 700\,ms p50 latency in all cases, and LLM-based inference, which frequently exceeds 1\,s even in zero-shot settings. This separation is particularly pronounced for Claude~4.5, where higher inference latency dominates regardless of prompting strategy.

\paragraph{Interpreting Pareto trade-offs.}
Although LLM prompting occasionally yields marginal improvements in macro-F1—most notably on IMDB and DBPedia—these gains are systematically offset by one to two orders of magnitude higher inference cost. Few-shot prompting further amplifies this effect by increasing input token counts, which directly translates into higher inference cost and shifts configurations deeper into a dominated region of the Pareto space.

As a result, encoder-based models remain Pareto-optimal across most realistic operating regimes, particularly when deployment constraints prioritize cost efficiency and low-latency inference.
Only in isolated accuracy-saturated scenarios do LLM-based configurations approach the Pareto frontier, and even then, they remain substantially more expensive than fine-tuned alternatives.

Taken together, the Pareto analysis reinforces the central conclusion of this work:
for fixed-label text classification tasks under production constraints, fine-tuned encoder models dominate large language model prompting across the vast majority of operational regimes.
Utility-based ranking then provides a principled mechanism for selecting among these non-dominated candidates when a single deployment decision must be made.

\section{Discussion}
\label{sec:discussion}

This study evaluates a critical architectural decision in modern NLP pipelines: whether to implement fixed-label text classification via fine-tuned encoder models or via prompting large language model (LLM) APIs. Using four benchmarks and a production-grade inference setup, we demonstrate that \emph{scale is not a substitute for specialization}. Our empirical results show that LLM prompting consistently yields inferior or only marginally superior predictive performance while introducing significantly higher latency—particularly in tail distributions—and incurring inference costs up to two orders of magnitude higher than specialized encoders at scale.

\subsection{Reassessing the Role of Scale in Fixed-Label Classification}

A persistent trend in contemporary NLP is the assumption that model scale is a universal proxy for performance. While foundation models excel in open-ended generation and zero-shot reasoning \cite{bommasani2021foundation, bubeck2023sparks}, our results suggest that for specialized, narrow-domain tasks with fixed label spaces, fine-tuned encoders represent a superior architectural paradigm.

This finding aligns with a nuance in scaling laws: when model capacity significantly exceeds task entropy, additional parameters yield marginal predictive gains that are systematically offset by increases in computational overhead, inference latency, and operational cost \cite{kaplan2020scaling, hoffmann2022training}. In this regime, the instruction-following capabilities of LLMs do not act as a substitute for the high-density task alignment achieved via supervised fine-tuning.

Empirically, this "tax of scale" is most evident in structured multi-class settings like AG News, where specialized encoders maintain a predictive advantage while operating with orders of magnitude higher efficiency. Even in near-saturated benchmarks such as DBPedia, encoders achieve the performance ceiling at a fraction of the cost, rendering LLM prompting a dominated strategy under any realistic production SLA.

\subsection{Operational Reality: Tail Latency, Controllability, and Systemic Risk}

In production environments, mean performance is seldom the primary driver of architectural choice; rather, it is the tail latency and the predictability of response times that dictate Service Level Agreement (SLA) compliance. Our empirical measurements reveal that encoder-based deployments operate within a tightly bounded latency regime. Conversely, LLM API inference exhibits substantial variance in the upper percentiles (p95 and p99) and a high Time-to-First-Token (TTFT), symptomatic of an external network-bound dependency. This variability is not merely a performance bottleneck; it fundamentally complicates downstream system design, including retry logic, throughput capacity planning, and timeout-triggered failures.

Beyond performance metrics, \emph{controllability} represents a critical, yet often overlooked, deployment property. Fine-tuned encoders are self-contained, versioned artifacts that enable deterministic regression testing and seamless rollback procedures. In contrast, API-based inference introduces a "black-box" dependency where model behavior, routing logic, and underlying infrastructure remain opaque. For organizations, this introduces significant risk: provider-side updates or load-induced fluctuations can degrade predictive performance or response stability without notice. For fixed-label classification, where consistency is paramount, the architectural autonomy provided by specialized encoders offers a level of auditability and risk mitigation that API-based solutions cannot currently match.

\subsection{Economic Perspective: The Trade-off between Convenience and Efficiency}

The economic findings of this study are unequivocal: even in scenarios where LLM predictive performance approaches that of fine-tuned encoders (e.g., IMDB), the cost-to-utility ratio remains decisively unfavorable. Few-shot prompting further exacerbates this disparity; by expanding the input context, it shifts model configurations deeper into a region of diminishing returns, where marginal F1 gains are systematically eclipsed by linear increases in token-based costs. At production scale, an architectural choice that introduces an order of magnitude higher latency and significantly greater operational expenditure is seldom justifiable for deterministic label mapping, unless the task necessitates generative capabilities that encoders cannot provide.

These results offer a critical lens through which to view the current ``LLM-first'' industrial trend. Our evidence suggests that the widespread adoption of LLM prompting for fixed-label classification is frequently driven by \emph{development convenience}---minimal pipeline overhead, rapid prototyping, and unified interfaces---rather than by empirical efficiency under production constraints. This indicates a strategic misalignment: LLMs are increasingly being deployed outside their comparative advantage. While their strength lies in high-entropy, open-ended reasoning, their application to learnable, closed-label tasks represents an inefficient allocation of both computational and financial resources.

\subsection{Interpretability and Auditability as First-Class Requirements}

Even when predictive performance is comparable, interpretability and auditability often dictate the viability of a system in regulated or high-stakes environments. Encoder-based classifiers are highly amenable to established attribution frameworks and diagnostic tooling, such as SHAP and Integrated Gradients \cite{lundberg2017shap, sundararajan2017ig}. These methods enable token-level explanations and feature importance analysis, which are critical for debugging, bias mitigation, and formal governance workflows.

In contrast, LLM prompting remains a comparatively opaque decision mechanism. Generative outputs are notably sensitive to minor variations in prompt phrasing and context window composition, leading to potential instability in the decision logic. Furthermore, explaining \emph{why} a specific label was assigned typically relies on post-hoc natural language justifications, which are not inherently faithful to the model's internal reasoning. For structured decision-making systems, this gap is not merely cosmetic; it represents a fundamental reliability and compliance constraint that favors the deterministic and auditable nature of specialized encoders.

\subsection{Environmental and Sustainability Implications}

Computational efficiency is not merely an economic concern; it is a fundamental pillar of sustainable AI development. Deploying high-parameter generative models for routine, fixed-label classification significantly amplifies energy demand relative to encoder-based inference. This disparity contributes to a disproportionate carbon footprint and increased energy consumption throughout the model's operational lifecycle \cite{strubell2019energy, schwartz2020greenai, patterson2021carbon}. 

Our results reinforce the principles of ``Green AI'' by demonstrating that specialized encoders achieve near-ceiling accuracy with a fraction of the computational footprint. At high request volumes, the cumulative energy savings of moving from API-based LLM prompting to specialized, stateless encoders are substantial. Consequently, we argue that for deterministic tasks where performance parity exists, the selection of the more efficient architecture is not only a matter of operational optimization but also a requirement for responsible and sustainable machine learning practices \cite{schwartz2020greenai}.

\subsection{LLMs as Knowledge Generators, Encoders as Decision Engines}

The empirical evidence supports a productive architectural distinction between model families. 
Large language models are particularly effective as \emph{knowledge generators}: they can assist in drafting labeling guidelines, proposing or refining label taxonomies, generating weak supervision signals, synthesizing training data, and supporting exploratory error analysis. 
In these roles, LLMs operate upstream of the final decision process, enriching the system’s knowledge artifacts rather than directly producing deterministic predictions.

Encoder-based models, by contrast, are better suited to function as \emph{decision engines} once the label space is fixed and the task requires fast, reproducible inference at scale. 
Fine-tuned encoders produce stable, versioned artifacts whose behavior can be validated through regression testing, monitored over time, and deployed under strict latency and cost constraints—properties that are central to production-grade decision systems.

This division of labor aligns with broader trends in modular NLP system design, including retrieval-augmented architectures, model distillation, and weak supervision pipelines, where generative models inform or guide compact task-specific learners rather than replacing them outright \cite{izacard2022atlas,raffel2020t5}. 
Viewed through this lens, the question is not whether LLMs or encoders are superior in isolation, but how each can be assigned to the role where its inductive biases and computational profile provide maximal system-level value.

\subsection{Threats to Validity and Limitations}
\label{sec:limitations}

\paragraph{Construct Validity.}
Our evaluation is constrained to fixed-label classification under deterministic decoding ($T=0$) and restricted output formats. While advanced prompting strategies---such as Chain-of-Thought (CoT) or self-consistency---might improve predictive performance in specific domains, they inherently increase the operational overhead in terms of token consumption and latency. Consequently, such methods would likely widen, rather than bridge, the efficiency gap identified in this study. Furthermore, while we prioritize macro-F1 to account for class imbalance, we acknowledge that other deployment-critical objectives, such as probability calibration and abstention logic, warrant further investigation.

\paragraph{Internal Validity.}
The latency and cost metrics for encoder models are tied to our specific Google Cloud Run configuration. Variables such as regional availability, concurrency limits, and cold-start behavior can influence absolute performance. We mitigate these factors by maintaining a unified infrastructure stack and reporting percentile-based statistics ($p95, p99$) to capture variance. For LLM APIs, the primary source of internal bias is provider-side stochasticity and transient network conditions. While our repeated-run protocol captures some of this variability, the underlying infrastructure remains an unobservable variable controlled by the vendors.

\paragraph{External Validity.}
The datasets selected (IMDB, SST-2, AG News, DBPedia) represent standard benchmarks for English text classification. The dominance of encoder specialization observed here may vary in domains characterized by high ambiguity, open-world reasoning, or long-form document adjudication where label discovery is required. However, for the majority of industrial use cases involving stable taxonomies and high-throughput requirements, our findings regarding the efficiency of specialized models are likely to remain robust.

\paragraph{Taxonomy and Data Quality.}
The ceiling for classification performance is fundamentally limited by label consistency and dataset cleanliness. In instances of taxonomy drift or underspecified label boundaries, switching to a larger model family is often a sub-optimal intervention. We argue that in such scenarios, resources are more effectively allocated toward improved annotation protocols and monitoring rather than increasing model scale.

\paragraph{Temporal Pricing and Vendor Lock-in.}
Economic calculations are based on pricing snapshots from January 2026. While the cost of inference is subject to market fluctuations, the magnitude of the observed disparities is large enough that moderate price reductions in LLM APIs are unlikely to reverse the qualitative conclusions of this work. Beyond unit cost, the structural risk of ongoing vendor dependence and variable expenditure remains a key differentiator compared to the predictable infrastructure costs of self-hosted encoder deployments.

\subsection{Summary}
Overall, the empirical evidence supports a clear deployment principle: for fixed-label text classification under production constraints, fine-tuned encoder models are the default \emph{right tool}. Across all benchmarks, they deliver equal or superior accuracy with one to two orders of magnitude lower inference cost and substantially lower, more predictable latency, making them better aligned with real-world SLA, budget, and reliability requirements. This outcome is consistent with prior analyses of scaling behavior in neural models, which show that increasing model size yields diminishing returns once task structure and label boundaries are well specified \cite{hernandez2021scaling}.

By contrast, large language models remain invaluable when flexibility, open-ended reasoning, or rapid schema evolution is required. However, our results indicate that their widespread industrial use as \emph{default} classifiers is rarely justified by accuracy--latency--cost trade-offs. Instead, this pattern appears to be driven primarily by development convenience, API accessibility, and organizational momentum rather than by measured system-level efficiency. Similar conclusions have been reported in recent empirical studies showing that general-purpose generative models are often outperformed by smaller, task-specialized systems once inference efficiency is taken into account \cite{strubell2019energy}.

From both an economic and sustainability perspective, deploying generative models for routine classification constitutes an avoidable inefficiency. Prior work has demonstrated that inference-time compute and serving overhead dominate the carbon and financial footprint of large-scale models in production environments, motivating a shift toward efficiency-aware model selection and deployment strategies \cite{patterson2021carbon}.

\section{Conclusion}

This study provides a comprehensive empirical and conceptual re-evaluation of the trade-offs between large generative models and fine-tuned encoder-based architectures for text classification. Across four benchmark datasets and through an extensive cost--latency--accuracy analysis, we demonstrate that fine-tuned models such as BERT, RoBERTa, and DistilBERT consistently outperform zero-shot and few-shot LLMs once deployment-relevant constraints are jointly considered. While the scaling of LLMs has yielded unprecedented general-purpose capabilities, these advantages remain largely underutilized in deterministic, fixed-label NLP tasks.

The central takeaway is clear: \textbf{large language models are not universal substitutes for task-specialized encoders}. Rather, the two model families should be understood as complementary components within a broader AI ecosystem—LLMs excelling at semantic abstraction and open-ended reasoning, and encoders providing precision, reproducibility, and efficiency.  
This finding challenges the prevailing “LLM-first” deployment mindset and motivates a shift toward \emph{task-fit model design}, where architectural choices are guided by empirical system-level trade-offs rather than by model scale alone \cite{bommasani2021foundation}.

From an engineering perspective, our results motivate hybrid pipelines in which large models operate upstream as meta-learners — generating weak supervision signals, candidate labels, or semantic enrichments — while compact encoders perform the final classification efficiently at scale. Such architectures naturally balance interpretability and throughput, reduce energy consumption, and integrate cleanly into auditable MLOps workflows \cite{zhou2024efficientllm}.

From an ethical and sustainability standpoint, this work reinforces the principle that progress in AI should not be measured by parameter count alone, but by \emph{effective intelligence per watt, per dollar, and per task}. Responsible NLP deployment requires explicit consideration of fairness, transparency, and environmental impact \cite{weidinger2022taxonomy, bender2021dangers}. In this respect, encoder-based models—open, controllable, and well understood—naturally align more closely with these values than opaque, usage-based LLM APIs.

Looking ahead, the evolution of applied NLP systems will hinge on combining reasoning power with deployment pragmatism.  
Promising directions for future work include:
\begin{itemize}
    \item efficient hybridization between encoder and decoder architectures;
    \item dynamic model selection via reinforcement learning and meta-optimization;
    \item life-cycle–aware evaluation frameworks that jointly account for accuracy, latency, cost, and environmental impact.
\end{itemize}

In summary, this benchmark is intended as a reusable \emph{knowledge artifact} for operational model selection rather than as a one-off performance comparison.  
By translating empirical measurements into decision criteria and minimal selection rules, we provide practical guidance for building cost-aware, auditable, and scalable text classification systems.  
More broadly, the results suggest that responsible NLP deployment requires allocating model capacity to the task context: encoders for stable, high-throughput decision making, and LLMs for knowledge generation and reasoning where their flexibility justifies the associated operational overhead.


\clearpage
\onecolumn
\appendix

\section{Cost Modeling Assumptions}
\label{app:cost_model}

To ensure transparent and reproducible cost comparisons, we formalize the inference-time cost models used throughout the paper.  
Costs are computed on a per-request basis and later scaled linearly to 1M requests, assuming steady-state operation.

\subsection{Encoder-based Models (BERT Family)}

For fine-tuned encoder models deployed as stateless services, inference cost is dominated by infrastructure usage during request processing.  
We approximate per-request cost using median end-to-end latency (p50) and allocated compute resources:

\[
\text{Cost}_{\text{Encoder}} =
Latency_{50}
\cdot
\left(
vCPU \cdot P_{vCPU}
\;+\;
GiB \cdot P_{GiB}
\right),
\]

where:
\begin{itemize}
  \item $Latency_{50}$ is the p50 inference latency per request (in seconds);
  \item $vCPU$ is the number of allocated virtual CPUs;
  \item $GiB$ is the allocated memory in gibibytes;
  \item $P_{vCPU}$ is the price per vCPU-second;
  \item $P_{GiB}$ is the price per GiB-second.
\end{itemize}

This formulation reflects serverless pricing models (e.g., Cloud Run), where cost scales linearly with execution time and allocated resources.  
Training cost is excluded, as the analysis focuses on steady-state inference economics.

\subsection{LLM-based Models (API Inference)}

For large language models accessed via external APIs, inference cost is token-based and independent of client-side compute.  
Per-request cost is computed as:

\[
\text{Cost}_{\text{LLM}} =
Tokens_{\text{in}} \cdot P_{\text{in}}
\;+\;
Tokens_{\text{out}} \cdot P_{\text{out}},
\]

where:
\begin{itemize}
  \item $Tokens_{\text{in}}$ and $Tokens_{\text{out}}$ denote average input and output tokens per request;
  \item $P_{\text{in}}$ and $P_{\text{out}}$ are the provider-specified prices per million input and output tokens.
\end{itemize}

Latency does not directly affect monetary cost in this regime, but it remains a critical deployment constraint and is therefore analyzed separately through p50/p95/p99 statistics and time-to-first-token (TTFT).

\subsection{Interpretation}

These cost models intentionally abstract away provider-specific discounts and batching optimizations in order to provide a conservative, comparable baseline.  
While absolute costs may vary across infrastructures and pricing revisions, the relative cost gaps observed between encoder-based and LLM-based inference remain robust under reasonable parameter variations.

\clearpage

\section{Fine-tuning Results and Epoch Selection}
\label{sec:finetuning}

All encoder-based models (BERT, DistilBERT, and RoBERTa) were fine-tuned for a total of four epochs across all datasets and random seeds. Rather than reporting the final epoch by default, we explicitly selected the optimal epoch for each experimental setting using a principled generalization-aware criterion. Under this procedure, the first epoch ($n=1$) emerged as the optimal operating point in every experiment, and we therefore report only results from this epoch throughout the paper.

Epoch selection is based on the following score:
\[
S = F_{\text{val}} - \lvert F_{\text{train}} - F_{\text{val}} \rvert,
\]
which simultaneously rewards strong validation performance and penalizes divergence between training and validation metrics. This criterion favors models that generalize well while discouraging overfitting, rather than those that merely optimize training performance.

Across all datasets, Epoch~1 consistently achieves the highest selection score.
Although later epochs often exhibit continued improvements in training F1, they also display increasing train--validation gaps, leading to lower selection scores despite comparable or marginally higher validation metrics.
This behavior signals the onset of overfitting and reflects the well-known tendency of large pretrained encoders to converge rapidly on downstream text classification tasks.

All reported results correspond to validation metrics obtained at Epoch~1 and are averaged across three independent random seeds. Fine-tuning is performed using only the training split, with the validation split used exclusively for checkpoint selection and performance reporting; the test split is not accessed during training or model selection. Mean and standard deviation are computed using an unbiased estimator (ddof~=~1). Model rankings, shown in parentheses, are based on mean validation F1, with higher values indicating better performance.

\subsection{IMDB}
\begin{table}[h]
\caption{IMDB — Fine-tuned encoder results on validation split across 3 seeds at Epoch~1 (mean$\pm$std). Higher is better. Rank shown in parentheses based on mean F1.}
\label{tab:finetune_imdb}
\centering
\begin{tabular}{lccccc}
\toprule
\textbf{Model} & \textbf{F1 (\%)} & \textbf{Precision (\%)} & \textbf{Recall (\%)} & \textbf{Accuracy (\%)} & \textbf{Train time} \\
\midrule
BERT       & 93.50$\pm$0.12 (2) & 93.55$\pm$0.10 & 93.51$\pm$0.12 & 93.51$\pm$0.12 & 8m 31s$\pm$6.7s \\
DistilBERT & 92.66$\pm$0.04 (3) & 92.69$\pm$0.07 & 92.66$\pm$0.04 & 92.66$\pm$0.04 & 4m 19s$\pm$0.0s \\
RoBERTa    & 95.14$\pm$0.16 (1) & 95.15$\pm$0.17 & 95.14$\pm$0.16 & 95.14$\pm$0.16 & 8m 18s$\pm$0.6s \\
\bottomrule
\end{tabular}
\end{table}

\subsection{SST-2}
\begin{table}[h]
\caption{SST-2 — Fine-tuned encoder results on validation split across 3 seeds at Epoch~1 (mean$\pm$std). Higher is better. Rank shown in parentheses based on mean F1.}
\label{tab:finetune_sst2}
\centering
\begin{tabular}{lccccc}
\toprule
\textbf{Model} & \textbf{F1 (\%)} & \textbf{Precision (\%)} & \textbf{Recall (\%)} & \textbf{Accuracy (\%)} & \textbf{Train time} \\
\midrule
BERT       & 94.66$\pm$0.26 (1) & 94.63$\pm$0.26 & 94.70$\pm$0.26 & 94.73$\pm$0.25 & 3m 08s$\pm$3.1s \\
DistilBERT & 93.72$\pm$0.09 (3) & 93.70$\pm$0.13 & 93.76$\pm$0.06 & 93.80$\pm$0.10 & 1m 40s$\pm$1.7s \\
RoBERTa    & 94.34$\pm$0.19 (2) & 94.38$\pm$0.15 & 94.31$\pm$0.21 & 94.43$\pm$0.18 & 3m 10s$\pm$2.0s \\
\bottomrule
\end{tabular}
\end{table}

\subsection{AG News}
\begin{table}[h]
\caption{AG News — Fine-tuned encoder results on validation split across 3 seeds at Epoch~1 (mean$\pm$std). Higher is better. Rank shown in parentheses based on mean F1.}
\label{tab:finetune_agnews}
\centering
\begin{tabular}{lccccc}
\toprule
\textbf{Model} & \textbf{F1 (\%)} & \textbf{Precision (\%)} & \textbf{Recall (\%)} & \textbf{Accuracy (\%)} & \textbf{Train time} \\
\midrule
BERT       & 94.31$\pm$0.07 (2) & 94.35$\pm$0.07 & 94.31$\pm$0.07 & 94.31$\pm$0.07 & 10m 01s$\pm$10.0s \\
DistilBERT & 94.28$\pm$0.06 (3) & 94.31$\pm$0.08 & 94.27$\pm$0.06 & 94.27$\pm$0.06 & 5m 23s$\pm$9.5s \\
RoBERTa    & 94.65$\pm$0.20 (1) & 94.67$\pm$0.21 & 94.65$\pm$0.20 & 94.65$\pm$0.20 & 9m 51s$\pm$2.6s \\
\bottomrule
\end{tabular}
\end{table}

\newpage

\subsection{DBpedia}
\begin{table}[h]
\caption{DBpedia — Fine-tuned encoder results on validation split across 3 seeds at Epoch~1 (mean$\pm$std). Higher is better. Rank shown in parentheses based on mean F1.}
\label{tab:finetune_dbpedia}
\centering
\begin{tabular}{lccccc}
\toprule
\textbf{Model} & \textbf{F1 (\%)} & \textbf{Precision (\%)} & \textbf{Recall (\%)} & \textbf{Accuracy (\%)} & \textbf{Train time} \\
\midrule
BERT       & 99.33$\pm$0.02 (1) & 99.33$\pm$0.02 & 99.33$\pm$0.02 & 99.33$\pm$0.02 & 49m 33s$\pm$18.4s \\
DistilBERT & 99.31$\pm$0.01 (2) & 99.31$\pm$0.01 & 99.31$\pm$0.01 & 99.31$\pm$0.01 & 26m 31s$\pm$8.3s \\
RoBERTa    & 99.27$\pm$0.02 (3) & 99.27$\pm$0.02 & 99.27$\pm$0.02 & 99.27$\pm$0.02 & 52m 48s$\pm$14.5s \\
\bottomrule
\end{tabular}
\end{table}

\clearpage

\section{Prompt Templates for LLM Evaluation}
\label{app:prompts}

To ensure full reproducibility of the LLM-based experiments, we report the exact prompt templates used for zero-shot (ZS) and few-shot (FS) evaluation across all datasets.  
All prompts enforce a constrained-output protocol: the model must return \emph{only} a single digit corresponding to the class label, with no additional text, punctuation, or explanation.  
All LLM runs use deterministic decoding ($T=0.0$) and the same label mappings described in the main text.

In the templates below, \texttt{\{text\}} denotes the dataset input string inserted verbatim at inference time.

\subsection{IMDB (Binary Sentiment)}

\paragraph{Zero-shot (ZS)}
\begin{verbatim}
Classify the sentiment of the following movie review from the IMDB dataset.
Respond only with a single digit: 0 if the sentiment is negative, 1 if the sentiment is positive.
Review: "{text}"
Label:
\end{verbatim}

\paragraph{Few-shot (FS)}
\begin{verbatim}
Classify the sentiment of the following movie review from the IMDB dataset.
Respond only with a single digit: 0 if the sentiment is negative, 1 if the sentiment is positive.
Return only the digit (no words, no punctuation).

Here are some examples:

Example 1:
Review: "Zentropa is the most original movie I've seen in years. If you like unique thrillers that are influenced by film 
noir, then this is just the right cure for all of those Hollywood summer blockbusters clogging the theaters these days. 
Von Trier's follow-ups like Breaking the Waves have gotten more acclaim, but this is really his best work. It is 
flashy without being distracting and offers the perfect combination of suspense and dark humor. It's too bad he 
decided handheld cameras were the wave of the future. It's hard to say who talked him away from the style he 
exhibits here, but it's everyone's loss that he went into his heavily theoretical dogma direction instead."
Label: 1

Example 2:
Review: "If only to avoid making this type of film in the future. This film is interesting as an experiment but tells 
no cogent story. One might feel virtuous for sitting thru it because it touches on so many IMPORTANT issues but
it does so without any discernable motive. The viewer comes away with no new perspectives (unless one comes up with 
one while one's mind wanders, as it will invariably do during this pointless film). One might better spend 
one's time staring out a window at a tree growing."
Label: 0

Now classify the following review:
Review: "{text}"
Label:
\end{verbatim}

\subsection{SST-2 (Binary Sentiment)}

\paragraph{Zero-shot (ZS)}
\begin{verbatim}
Classify the sentiment of the following movie sentence from the Stanford Sentiment Treebank v2 (SST-2) dataset.
Respond only with a single digit: 0 if the sentiment is negative, 1 if the sentiment is positive.
Sentence: "{text}"
Label:
\end{verbatim}

\paragraph{Few-shot (FS)}
\begin{verbatim}
Classify the sentiment of the following movie sentence from the Stanford Sentiment Treebank v2 (SST-2) dataset.
Respond only with a single digit: 0 if the sentiment is negative, 1 if the sentiment is positive.
Return only the digit (no words, no punctuation).

Here are some examples:

Example 1:
Sentence: "of those unassuming films that sneaks up on you and stays with you long after you have left the theater"
Label: 1

Example 2:
Sentence: "we do n't even like their characters ."
Label: 0

Now classify the following sentence:
Sentence: "{text}"
Label:
\end{verbatim}

\subsection{AG News (4-way Topic Classification)}

\paragraph{Zero-shot (ZS)}
\begin{verbatim}
Classify the topic of the following news article from the AG News dataset.
Respond only with a single digit according to the category:
0 = World
1 = Sports
2 = Business
3 = Sci/Tech
Article: "{text}"
Label:
\end{verbatim}

\paragraph{Few-shot (FS)}
\begin{verbatim}
Classify the topic of the following news article from the AG News dataset.
Respond only with a single digit according to the category:
0 = World
1 = Sports
2 = Business
3 = Sci/Tech
Return only the digit (no words, no punctuation).

Here are some examples:

Example 1:
Article: "Kerry leading Bush in key swing states (AFP) AFP - Although polls show the US presidential race a virtual 
dead heat, Democrat John Kerry appears to be gaining an edge over George W. Bush among the key states that could decide 
the outcome."
Label: 0

Example 2:
Article: "Colander Misses Chance to Emulate Jones  ATHENS (Reuters) - But for a decision that enraged her coach, LaTasha
Colander might have been the Marion Jones of the Athens Olympics."
Label: 1

Example 3:
Article: "Oil and Economy Cloud Stocks' Outlook  NEW YORK (Reuters) - Soaring crude prices plus worries about the economy 
and the outlook for earnings are expected to hang over the stock market next week during the depth of the summer doldrums."
Label: 2

Example 4:
Article: "Apple to open second Japanese retail store this month (MacCentral) MacCentral - Apple Computer Inc. will open 
its second Japanese retail store later this month in the western Japanese city of Osaka, it said Thursday."
Label: 3

Now classify the following article:
Article: "{text}"
Label:
\end{verbatim}

\subsection{DBPedia (14-way Ontology Classification)}

\paragraph{Zero-shot (ZS)}
\begin{verbatim}
Classify the topic of the following text from the DBpedia Ontology dataset.
Respond ONLY with a single digit (0–13) according to the category:
0 = Company
1 = EducationalInstitution
2 = Artist
3 = Athlete
4 = OfficeHolder
5 = MeanOfTransportation
6 = Building
7 = NaturalPlace
8 = Village
9 = Animal
10 = Plant
11 = Album
12 = Film
13 = WrittenWork
Do not include any explanation or text, only the digit.
Text: "{text}"
Label:
\end{verbatim}

\paragraph{Few-shot (FS)}

\begin{verbatim}
Classify the topic of the following text from the DBpedia Ontology dataset.
Respond only with a single digit (0-13) according to the category:

0 = Company
1 = EducationalInstitution
2 = Artist
3 = Athlete
4 = OfficeHolder
5 = MeanOfTransportation
6 = Building
7 = NaturalPlace
8 = Village
9 = Animal
10 = Plant
11 = Album
12 = Film
13 = WrittenWork

Return only the digit (no words, no punctuation).

Here are some examples:

Example 1:
Text:
"Angstrem Group (Russian: OAO "Angstrem" named after angstrom)
is a group of Russian companies, one of the largest manufacturers
of integrated circuits in Eastern Europe."
Label: 0

Example 2:
Text:
"Kirk Balk Community College is a state school in Barnsley,
South Yorkshire, England. It is a technology specialist college."
Label: 1

Example 3:
Text:
"Brian Robert Setzer (born April 10 1959) is an American guitarist,
singer and songwriter. He first found widespread success in the
early 1980s with the rockabilly revival group Stray Cats and later
with The Brian Setzer Orchestra."
Label: 2

Example 4:
Text:
"Colin Henry Turkington (born 21 March 1982) is a Northern Irish
auto racing driver and 2009 British Touring Car Champion."
Label: 3

Example 5:
Text:
"Maria Antonieta de Bogran (born 13 July 1955) is the former
1st Vice President of Honduras. She was a candidate for Vice
President in the 2009 Honduran elections and served as national
chairperson of the National Party."
Label: 4

Example 6:
Text:
"The South African Class 8E of 1983 is a South African electric
locomotive used in shunting service."
Label: 5

Example 7:
Text:
"The Lefferts Historic House located in Brooklyn's Prospect Park
is a historic home and museum built circa 1783."
Label: 6

Example 8:
Text:
"Ostensjovannet is a lake in Oslo, Norway, known for its birdlife
and wildlife preserve status."
Label: 7

Example 9:
Text:
"Malgammana is a village located in the Central Province of
Sri Lanka."
Label: 8

Example 10:
Text:
"The Tawny Coster (Acraea terpsicore) is a butterfly species
belonging to the Nymphalidae family."
Label: 9

Example 11:
Text:
"Cupressus abramsiana is a species of cypress endemic to the
Santa Cruz Mountains in California."
Label: 10

Example 12:
Text:
"Kim Appleby is the solo debut album by British singer
Kim Appleby released in 1990."
Label: 11

Example 13:
Text:
"Yitzhak Rabin: a Biography is a 2004 documentary film about the
life of Israeli Prime Minister Yitzhak Rabin."
Label: 12

Example 14:
Text:
"The Cameo Murders is a book by Barry Shortall detailing a
1949 murder case in Liverpool."
Label: 13

Now classify the following text:

Text: "{text}"
Label:
\end{verbatim}

\twocolumn
\clearpage

\bibliographystyle{ACM-Reference-Format}
\bibliography{references}

\end{document}